\documentclass[10pt,preprint]{article}
\PassOptionsToPackage{table,xcdraw}{xcolor}  
\let\origaddcontentsline\addcontentsline

\usepackage{rlc}

\let\addcontentsline\origaddcontentsline

\usepackage{hyperref}
\hypersetup{
    colorlinks, linkcolor={dark-blue},
    citecolor={dark-blue}, urlcolor={dark-blue}
}

\usepackage{etoolbox}
\pretocmd{\section}{\phantomsection}{}{}
\pretocmd{\subsection}{\phantomsection}{}{}

\usepackage{authblk} 
\usepackage{amssymb}            
\usepackage{mathtools}          
\usepackage{mathrsfs}           
\mathtoolsset{showonlyrefs}     
\usepackage{graphicx}           
\usepackage{subcaption}         
\usepackage[space]{grffile}     
\usepackage{url}                
\usepackage{booktabs}       
\usepackage{amsfonts}       
\usepackage{nicefrac}       
\usepackage{microtype}      

\usepackage{graphicx}
\usepackage{natbib}
\usepackage{doi}
\usepackage{svg}
\usepackage{tikz}
\usepackage[table]{xcolor}
\usetikzlibrary{trees, shapes, arrows, positioning, calc}
\usepackage{array, makecell, booktabs}
\newcolumntype{P}[1]{>{\raggedright\arraybackslash}p{#1}}
\definecolor{greyboxbg}{RGB}{221,221,221}
\definecolor{blueboxbg}{RGB}{240,240,240}
\definecolor{orangeboxbg}{RGB}{200,255,200}
\definecolor{greenboxbg}{RGB}{142,207,201}
\usepackage{multirow} 

\title{\vspace{+3cm}A Survey of AI Agent Protocols\vspace{0cm}}

\author{\\
\name{Yingxuan Yang}, \name{Huacan Chai}, \name{Yuanyi Song}, \name{Siyuan Qi}, \\
\name{Muning Wen}, \name{Ning Li}, \name{Junwei Liao}, \name{Haoyi Hu}, \name{Jianghao Lin\thanks{Corresponding author.}}, \\
\name{Gaowei Chang}$^\dag$, \name{Weiwen Liu}, \name{Ying Wen}, \name{Yong Yu}, \name{Weinan Zhang} \vspace{+0.15cm}\\
Shanghai Jiao Tong University, $^\dag$\text{ANP Community} \\
\texttt{\{zoeyyx, chiangel, wnzhang\}@sjtu.edu.cn}
}

\begin{document}
\maketitle

\begin{abstract}
	The rapid development of large language models (LLMs) has led to the widespread deployment of LLM agents across diverse industries, including customer service, content generation, data analysis, and even healthcare. However, as more LLM agents are deployed, a major issue has emerged: there is no standard way for these agents to communicate with external tools or data sources. This lack of standardized protocols makes it difficult for agents to work together or scale effectively, and it limits their ability to tackle complex, real-world tasks. A unified communication protocol for LLM agents could change this. It would allow agents and tools to interact more smoothly, encourage collaboration, and triggering the formation of collective intelligence.
    In this paper, we provide the first comprehensive analysis of existing agent protocols, proposing a systematic two-dimensional classification that differentiates context-oriented versus inter-agent protocols and general-purpose versus domain-specific protocols. Additionally, we conduct a comparative performance analysis of these protocols across key dimensions such as security, scalability, and latency. 
    Finally, we explore the future landscape of agent protocols by identifying critical research directions and characteristics necessary for next-generation protocols. These characteristics include adaptability, privacy preservation, and group-based interaction, as well as trends toward layered architectures and collective intelligence infrastructures.
    We expect this work to serve as a practical reference for both researchers and engineers seeking to design, evaluate, or integrate robust communication infrastructures for intelligent agents. Ongoing updates and the latest developments will be maintained at \url{https://github.com/zoe-yyx/Awesome-AIAgent-Protocol}.
\end{abstract}

\hspace{32pt} \textbf{Key Words:} AI Agent Protocol, AI Agent, Agent Protocol Evaluation, LLMs

\newpage
\tableofcontents

\section{Introduction}
With the rapid advancement of large language models (LLMs), LLM agents\footnote{For presentation brevity, in this paper, the multi-modal LLM concept \citep{caffagni2024r} is merged into the LLM concept.} are increasingly being deployed across various industries, including automated customer service, content creation, data analysis, and medical assistance~\citep{openai2024gpt4technicalreport,gottweis2025aicoscientist,yang2025whosmvpgametheoreticevaluation,Guo2024LargeLM,zhou2024tradenhancingllmagents}, transforming our daily work and life. To fully exploit the potential of agents, many architectures have emerged to facilitate communication between agents and external entities. These entities include resources not directly controlled by agents, such as various data sources and tools, as well as other online agents.

However, as the scope of application scenarios expands and agents from different vendors with different structures emerge, the interaction rules between agents and entities have grown complex. A critical bottleneck in this evolution is the absence of standardized protocols. This deficiency hinders agent interoperability with aforementioned resources~\citep{Qu_2025, patil2023gorillalargelanguagemodel, liu2024apigenautomatedpipelinegenerating}, limiting their capability to leverage external functionalities. In addition, the lack of standardized protocols prevents seamless collaboration between agents from different providers or architectural backgrounds, thus limiting the scalability of agent networks. Ultimately, the ability of agents to solve more complex real-world problems is thereby limited.

These challenges echo a pivotal moment in computing history when the early Internet was fragmented by incompatible systems and limited connectivity. Nowadays, the landscape of LLM agents suffers from similar isolation. The revolutionary impact of TCP/IP and HTTP protocols didn't merely solve technical problems—they unleashed an unprecedented era of global connectivity, innovation, and value creation that transformed human society.
Similarly, a unified protocol for agent systems wouldn't just address current interoperability issues—it would create something far more transformative: a connected network of intelligence~\citep{Multi-Agent-as-a-Service, yang2024llmbasedmultiagentsystemstechniques,chen2024internet,yang2025agentnetdecentralizedevolutionarycoordination}. Such standardization would enable different forms of intelligence to flow between systems—where tools with embedded intelligence could seamlessly interact with specialized agents, combining their capabilities to create emergent forms of collective intelligence greater than any individual component. This intelligence network would break down the artificial barriers between "tool intelligence" and "agent intelligence", allowing them to merge, amplify, and complement each other dynamically. Specialized agents could form temporary coalitions to solve complex problems, intelligent tools could extend the capabilities of multiple agents simultaneously, and entirely new cognitive architectures could emerge from these standardized interactions. The result wouldn't merely be more efficient automation but a fundamentally new paradigm of distributed, collaborative intelligence that could address challenges beyond the reach of today's isolated systems.

To address the aforementioned limitations, existing work continues to advance the standardization of protocols. For instance, in agent-to-resource communication, Anthropic has introduced the Model Context Protocol (MCP)~\citep{anthropic2025}, which standardizes context acquisition between LLM agents and external resources. MCP greatly enhances agents' ability to communicate with external data and tools, effectively acting as an "external brain" to augment agent knowledge and tackle complex real-world problems more efficiently. Similarly, protocols like Agent Network Protocol (ANP) \citep{anp2024} and Agent-to-Agent (A2A) \citep{a2a2025} facilitate collaboration among agents from diverse providers and structures in multi-agent scenarios. However, despite their rapid development, the lack of a detailed analysis and survey of agent protocols results in users and developers encountering difficulties when navigating the extensive agent protocols in practice. Among the most pressing concerns for users and developers are the analysis and classification of the similarities and differences between protocols, as well as the comparison of their various performance characteristics.

This survey provides the first comprehensive analysis of existing agent protocols. Through a detailed investigation of protocols, we present a systematic classification of agent protocols for the first time, offering a clear framework for the numerous available protocols, thereby assisting users and developers in selecting the most suitable protocols for specific scenarios. Furthermore, we undertake a comparative analysis of the performance of various protocols across multiple key dimensions, including security, scalability, and latency, providing valuable insights for future research and practical applications of agent protocols. 
Finally, we explore the future landscape of LLM agent protocol, outlining major research directions and identifying the characteristics that next-generation protocols should embody to support evolving agent ecosystems, such as adaptability, privacy preservation, and group-based interaction.

\begin{figure*}[t]
    \centering
    \includegraphics[width=0.9\linewidth]{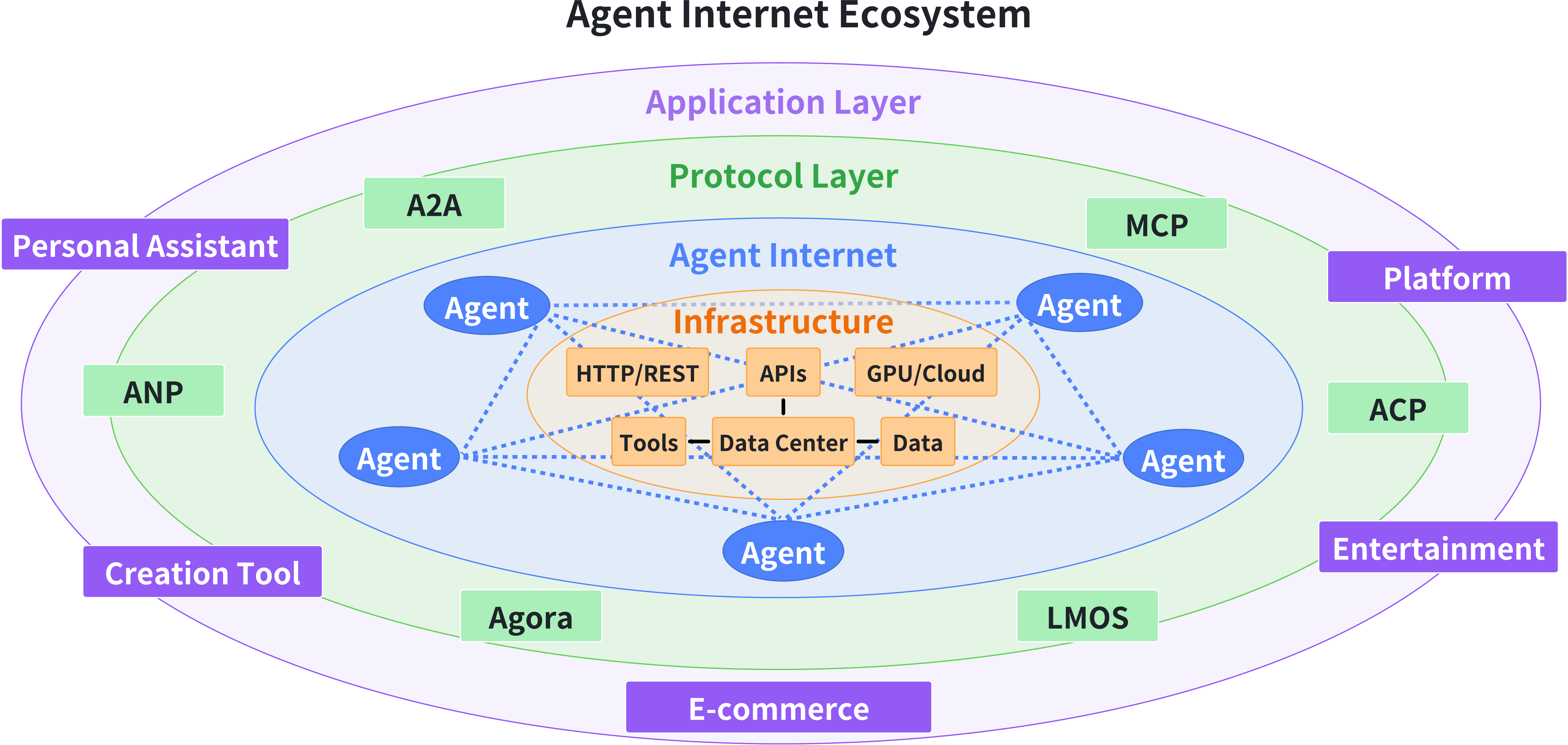}
    \caption{A layered architecture of the Agent Internet Ecosystem.}
    \label{fig:ecosystem}
\end{figure*}

In summary, our research makes several significant contributions to the field:
\begin{itemize}
\item We propose the first systematic, two-dimensional classification of agent protocols—distinguishing context-oriented vs. inter-agent protocols and general-purpose vs. domain-specific protocols—to provide a clear organizational framework. 
\item We conduct a qualitative analysis of current agent protocols across key dimensions such as efficiency, scalability, security, and reliability, revealing their relative strengths and limitations in different application environments.
\item We offer a forward-looking perspective on the evolution of agent protocols, identifying short-, mid-, and long-term trends, including the shift toward evolvable, privacy-aware, and group-coordinated protocols, as well as the emergence of layered architectures and collective intelligence infrastructures.
\end{itemize}

\begin{figure*}[t]
    \centering
    \includegraphics[width=\linewidth]{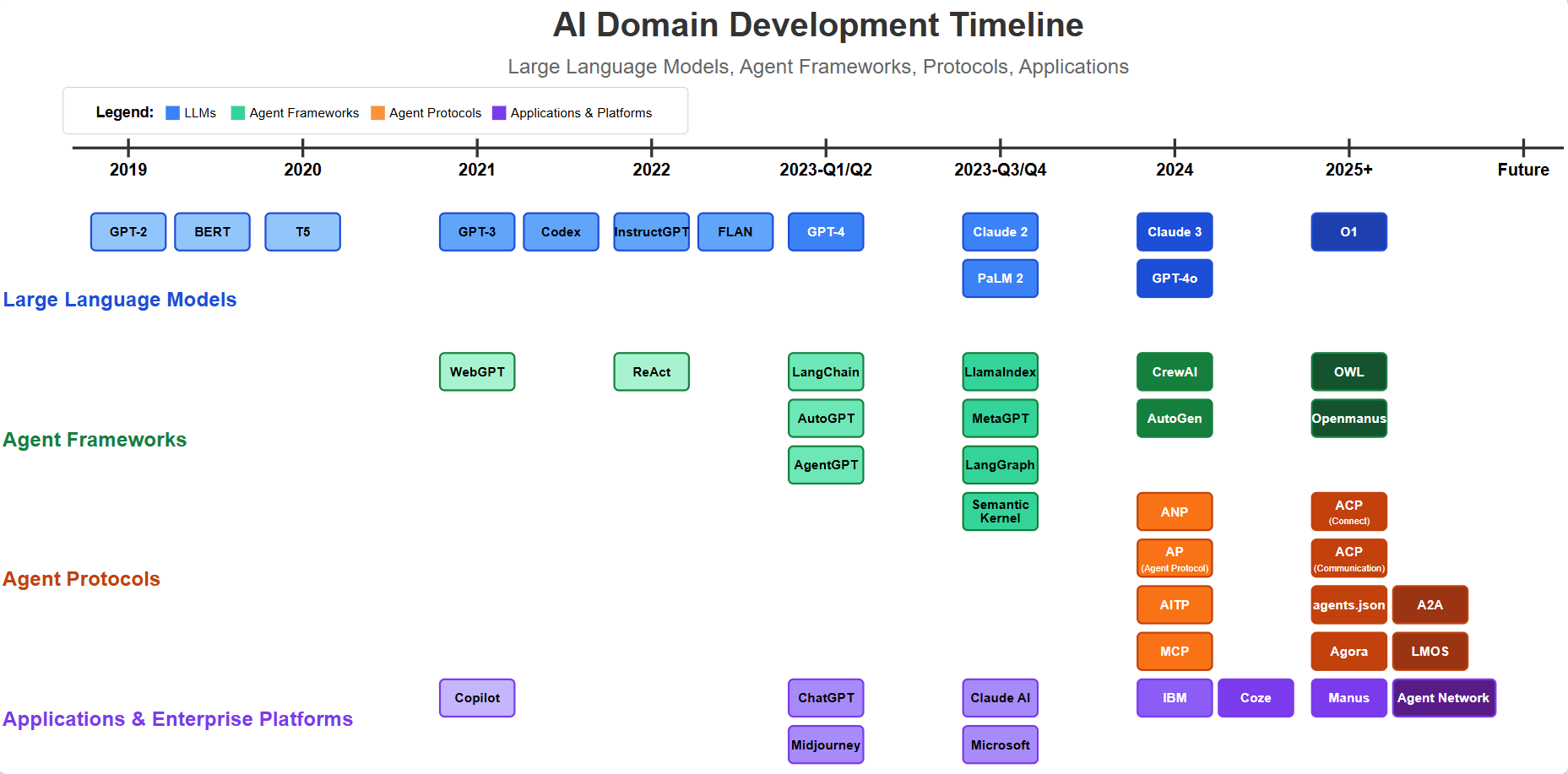}
    \caption{A glance at the development of agent protocols.}
    \label{fig:development}
\end{figure*}

\section{Preliminaries}
In this section, we present foundational concepts essential to understanding the subsequent survey and analyses. We first define LLM agents and discuss their key characteristics. Subsequently, we introduce the concept of agent protocols and their fundamental roles within LLM ecosystems.

\subsection{Definition and Characteristics of LLM agents}

LLM agents signify a notable advancement in artificial intelligence by integrating the sophisticated linguistic processing capabilities inherent in large language models with autonomous decision-making frameworks~\citep{yao2022react,autogpt,hong2024metagpt}. Specifically, these agents are advanced systems capable of generating complex textual outputs requiring sequential reasoning. They demonstrate capabilities such as forward-looking planning, maintaining contextual memory of past interactions, and employing external tools to dynamically adapt responses according to situational demands and desired communication styles.

What distinguishes LLM agents from standard Large Language Models is their architectural composition and operational capabilities. While LLMs primarily focus on text generation based on input prompts, agents are designed to function autonomously within real-world environments. The core architecture of an LLM agent typically consists of:
\begin{itemize}
    \item \textbf{Foundation Model}: 
    The core of an LLM-based agent is its foundation model~\citep{zhao2025surveylargelanguagemodels,Yin_2024}, typically a large language model or a multimodal large model, which provides essential capabilities for reasoning, understanding language, and interpreting multimodal information.
    \item \textbf{Memory Systems}: LLM agents implement both short-term and long-term memory components to maintain context across interactions and store relevant information for future use~\citep{zhang2024surveymemorymechanismlarge,yang2025agentnetdecentralizedevolutionarycoordination}. This dual memory system allows agents to maintain conversation continuity while building knowledge over time.
    \item \textbf{Planning}: Planning is a fundamental aspect of agent research~\citep{}, enabling agents to break down complex tasks into smaller, manageable subtasks. Such planning mechanisms facilitate strategic problem-solving and enhance the interpretability and transparency of the agent's decision-making processes.
    \item \textbf{Tool-Using}: 
    Although LLMs inherently face limitations in mathematical reasoning, logical operations, and knowledge beyond their trained corpus, agents overcome these constraints by integrating external tools and APIs\cite{wang2023toolllm,schick2023toolformerlanguagemodelsteach,Qu_2025,liu2024apigenautomatedpipelinegenerating}. Through systematic tool invocation, agents significantly extend their functionality and accuracy in responding to complex queries.
    \item \textbf{Action Execution}: The ability to interact with their environment by executing actions\cite{liu2023agentbench,yang2025whosmvpgametheoreticevaluation}, whether through API calls, database queries, or interaction with external systems.
\end{itemize}

The architectural components described above represent the foundational structure of modern LLM agents.  Building upon this architecture, recent advances in both academic research and industrial applications have significantly expanded agent capabilities and deployment scenarios.

\paragraph{Industrial Advancements}
In the industrial landscape, major technology companies have developed increasingly sophisticated agent platforms that leverage these architectural principles while adding enterprise-scale capabilities.   Microsoft has positioned itself as a leader by creating a comprehensive agent ecosystem that integrates with over 1,400 enterprise systems and allows the use of multiple LLM options beyond their OpenAI partnership~\citep{microsoft2024agents}.   Their autonomous agents can now handle complex workflows with minimal human supervision, particularly in areas such as sales automation, customer service, and business process optimization.
Similarly, IBM has embraced agent technology with their research indicating nearly universal adoption intentions among enterprise AI developers~\citep{ibm2024agent}.   Their focus on distinguishing between simple function-calling systems and truly autonomous agents with robust reasoning capabilities reflects the industry's growing recognition that advanced planning and reasoning components are essential for meaningful agent applications.
The democratization of agent development has accelerated through platforms like Coze~\citep{trustedby2024coze}, which enables non-technical users to build and deploy sophisticated agents across various communication channels.   This trend toward accessible development tools has broadened agent adoption across sectors, from specialized enterprise applications to consumer-facing implementations.

\paragraph{Academic Research Directions}
Academic research has increasingly focused on enhancing agent reasoning capabilities through specialized models designed specifically for complex analytical tasks. The development of reasoning-focused or o1-like models \citep{jaech2024openai} represents a significant advancement in enabling agents to handle intricate problem-solving scenarios that require multi-step logical processes.
Another key research direction involves multi-agent architectures where multiple specialized agents collaborate to accomplish complex tasks~\citep{yang2024llmbasedmultiagentsystemstechniques,Guo2024LargeLM,yang2025agentnetdecentralizedevolutionarycoordination,Multi-Agent-as-a-Service}. These systems distribute cognitive load across multiple agents, each optimized for specific subtasks, and have demonstrated superior performance in handling complex, open-ended problems compared to single-agent approaches.


\paragraph{Emerging Implementation Frameworks}
The practical implementation of agent systems has been facilitated by specialized frameworks that provide developers with pre-built components for agent construction. LangChain and its extension LangGraph have become industry standards for agent development~\citep{langchain2024langgraph}, offering modular architectures that support sophisticated reasoning, planning, and multi-agent coordination.
Microsoft's Semantic Kernel framework has focused on bridging traditional software development with AI capabilities~\citep{microsoft2024semantic}, making it easier to integrate agent functionality into existing enterprise systems without complete architectural overhauls. This integration-focused approach has been particularly valuable for enterprises seeking to enhance existing workflows rather than replace them.

These advancements collectively demonstrate the rapid evolution of LLM agents from experimental concepts to practical, value-generating systems deployed across diverse application domains. As the technology continues to mature, the integration of more sophisticated reasoning, planning, and action execution capabilities promises to further expand the role of autonomous agents in both enterprise and consumer contexts.

\definecolor{rowshade}{RGB}{240,240,240} 
\begin{table}[t]
\centering
\renewcommand{\arraystretch}{1.3} 
\caption{Comparison of the properties of different interaction manners for agents.}
\resizebox{\textwidth}{!}{
\begin{tabular}{p{2.5cm}|p{4.5cm}|c|c|c|c}
\toprule[1.1pt] 
\rowcolor{rowshade}
\textbf{Manner} & \textbf{Scenarios} & \textbf{Efficiency} & \textbf{Operation Range} & \textbf{Standardized} & \textbf{AI-Native} \\
\hline
\textcolor{black}{API} & Server-to-server integration & $\checkmark\checkmark$ & $\times$ & $\times$ & $\times$ \\
\hline
\textcolor{black}{GUI} & Computer/ Mobile Use & $\times$ & $\checkmark$ & $\checkmark$ & $\times$ \\
\hline
\textcolor{black}{XML} & Browser Use & $\times$ & $\checkmark$ & $\times$ & $\times$ \\
\hline
\textcolor{blue}{\textbf{Protocol}} & Agent Interaction & $\checkmark\checkmark$ & $\checkmark\checkmark$ & $\checkmark\checkmark$ & $\checkmark\checkmark$ \\
\bottomrule[1.1pt]
\end{tabular}}
\label{tab:interaction-comparison}
\end{table}

\subsection{Definition and Developments of Agent Protocols}

Agent protocols are standardized frameworks that define the rules, formats, and procedures for structured communication among agents and between agents and external systems. Compared to traditional interaction mechanisms—such as APIs, graphical user interfaces (GUIs), or XML-based interactions—protocols exhibit significant advantages, as summarized in Table~\ref{tab:interaction-comparison}. Unlike APIs, which are efficient but often lack operational flexibility and standardization, or GUIs, which provide user-friendly standardized interfaces but are limited in efficiency and not inherently AI-native, protocols combine the benefits of high efficiency, extensive operational scope, robust standardization, and native compatibility with AI systems. XML-based methods, primarily intended for browser-based interactions, similarly lack both efficiency and comprehensive standardization. Furthermore, a significant number of AI assistants focusing on browser-usage typically depend on HTML and other programming languages and analogous technologies for interactions between LLMs and websites. Nevertheless, this methodology is deficient in terms of flexibility and complexity, which hinders its applicability to alternative scenarios. Thus, agent protocols stand out as uniquely capable of supporting complex, dynamic, and scalable interactions within diverse agent ecosystems, making them the preferred approach for agent-based system communications.

Protocols serve as a foundational grammar enabling coherent information exchange, allowing heterogeneous agent systems to collaborate seamlessly regardless of their internal architectural differences. The primary value of these protocols lies in enabling interoperability, ensuring standardized interactions, and allowing agents to easily integrate and extend their capabilities by incorporating new tools, APIs, or services. Moreover, standardized protocols provide inherent mechanisms for maintaining security and governance, thereby managing agent behaviors within clearly defined and safe operational parameters. By abstracting away the complexities of interaction logic, protocols significantly reduce agent development complexity, empowering developers to concentrate efforts on enhancing core agent functionalities. Perhaps most transformatively, protocols enable collective intelligence to emerge when specialized agents form temporary coalitions to solve complex problems. By sharing insights and coordinating actions through standardized communication channels, distributed agent systems can achieve results impossible for monolithic architectures, enabling entirely new cognitive architectures that distribute reasoning across multiple specialized systems.

The current agent protocol landscape encompasses various strategic paradigms. Model-centric protocols, exemplified by Anthropic's Model Context Protocol (MCP)~\citep{anthropic2025}, seek ecosystem influence and asset control by major technology providers. Enterprise-focused protocols such as Agent-to-Agent (A2A)~\citep{a2a2025} prioritize integration, security, and governance within internal corporate environments. Meanwhile, open network protocols like the Agent Network Protocol (ANP)~\citep{anp2024} represent a decentralized vision, aiming to establish an open agent internet that encourages widespread agent interoperability regardless of provider or technology stack. These developments illustrate the critical role of protocols in advancing agent-based collaborative intelligence across diverse application domains.

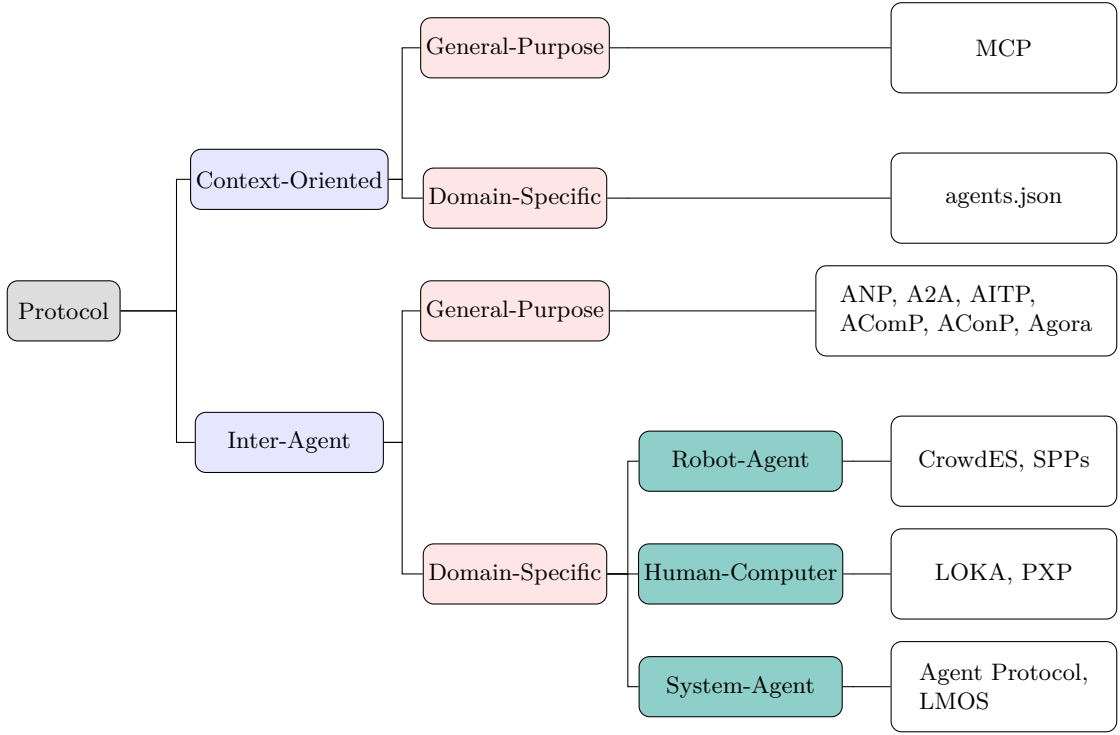
\begin{figure}[h]
\centering
\begin{tikzpicture}[
    level 1/.style={sibling distance=3.5cm, level distance=3cm},
    level 2/.style={sibling distance=3.5cm,level distance=3cm},
    level 3/.style={sibling distance=1.5cm, level distance=3cm},
    level 4/.style={sibling distance=2cm},
    every node/.style={align=center, font=\small},
    box/.style={rectangle, draw=black, fill=blue!10, rounded corners, inner sep=3pt, minimum width=1.5cm, minimum height=0.8cm},
    root/.style={rectangle, draw=black, fill=greyboxbg, rounded corners, inner sep=3pt, minimum width=1.5cm, minimum height=0.8cm},
    category1/.style={rectangle, draw=black, fill=blue!10, rounded corners, inner sep=2pt, minimum width=2.5cm, minimum height=0.8cm},
    category2/.style={rectangle, draw=black, fill=red!10, rounded corners, inner sep=2pt, minimum width=1.5cm, minimum height=0.8cm},
    category3/.style={rectangle, draw=black, fill=greenboxbg, rounded corners, inner sep=2pt, minimum width=2.7cm, minimum height=0.8cm},
    protocol/.style={rectangle, draw=black, fill=white, rounded corners, inner sep=2pt, minimum width=3cm, minimum height=1.2cm, align=left},
    category/.style={rectangle, draw=black, fill=red!10, rounded corners, inner sep=2pt},
    func/.style={rectangle, draw=black, fill=red!10, rounded corners, inner sep=2pt},
    grow=right,
    edge from parent fork right
]

\node[root] {Protocol}
    child {node[category1] {Inter-Agent}
        child {node[category2] {Domain-Specific}
            child {node[category3] {System-Agent\\ (Sec. 3.2.2.3)}
                child {node[protocol, xshift=0.5cm] {Agent Protocol,\\LMOS}}
            }
            child {node[category3] {Robot-Agent\\ (Sec. 3.2.2.2)}
                child {node[protocol, xshift=0.5cm] {CrowdES, SPPs}}
            }
            child {node[category3] {Human-Computer\\ (Sec. 3.2.2.1)}
                child {node[protocol, xshift=0.5cm] {LOKA, PXP, WAP}}
            }
        }
        child {node[category2] {General-Purpose}
            child {node[protocol, xshift=3cm, minimum width=4cm] {ANP, A2A, AITP, AComP,\\ AConP, Agora, Coral}}
        }
    }
    child {node[category1] {Context-Oriented}
        child [sibling distance=0.5cm] {node[category2] {Domain-Specific}
            child {node[protocol, xshift=3.5cm] {agents.json}}
        }   
        child {node[category2] {General-Purpose}
                child {node[protocol, xshift=3.5cm] {MCP}}
        }
        }
;
\end{tikzpicture}
\caption{Classification of various agent protocols from two dimensions, i.e., \colorbox{blue!10}{object orientation} and \colorbox{red!10}{application scenario}. Please refer to Table \ref{tab:overview-of-protocols} for more information. To distinguish between the two ACPs — the Agent Connect Protocol and the Agent Communication Protocol — we will refer to them as AconP and AcomP, respectively.}
\label{fig:classification-figure}
\end{figure}

\section{Protocol Taxonomy}

In response to the rapidly evolving demands of LLM agents, a variety of agent protocols have emerged. However, existing studies are deficient in the systematic classification of these protocols. To address this gap, we propose a two-dimensional classification framework for agent protocols illustrated in Figure~\ref{fig:classification-figure}. On the first dimension—\textbf{object orientation}—protocols are divided into context-oriented and inter-agent types; on the second dimension—\textbf{application scenario}—they are further categorized as general-purpose or domain-specific.

\subsection{Context-Oriented Protocols}
Despite the advanced language understanding and reasoning capabilities of LLMs, LLM agents cannot solely rely on the inherent knowledge of LLMs to respond to complex queries or intents. Instead, to get necessary context to achieve goals, LLM agents usually need to autonomously determine when and which external tools to invoke, and execute actions through these tools~\citep{liu2025advanceschallengesfoundationagents}. For example, consider a scenario in which a user poses a question about the weather at a particular date and location. In such an instance, LLM agents will independently decide to consult a real-world weather API to retrieve the pertinent data to obtain the necessary context and answer this question. In the early stages of development, the tool usage capability of LLM agents was typically fine-tuned through formatted function-calling datasets~\citep{Qu_2025, schick2023toolformerlanguagemodelsteach, liu2024apigenautomatedpipelinegenerating}. While this approach can quickly enhance the ability of LLM agents to invoke functions and require contexts, it faces several challenges due to the lack of a standardized and unified context-oriented protocol. 

However, the absence of standardized protocols in the LLM ecosystem has led to significant fragmentation in both tool invocations and interfaces. LLM providers often implement proprietary standards for tool usage, consequently leading to varying prompt formats across base models. Similarly, data, tool and service providers implement their own invocation interfaces, further exacerbating incompatibility. This fragmentation increases the burden on users and developers, requiring prompt-level customizations and management of diverse specifications, ultimately hindering interoperability, increasing system complexity, and raising development and maintenance costs.

In response to these challenges, several context-oriented agent protocols have been proposed. By providing a standardized method for context acquisition, these protocols can reduce fragmentation in context exchange between agents and context providers. Based on application scenarios, context-oriented agent protocols can be categorized as general-purpose or domain-specific. General-purpose protocols aim to support a wide range of agents and context providers through a unified interface, while domain-specific protocols focus on specialized optimization for particular use cases.

\definecolor{rowshade}{RGB}{240,240,240} 
\begin{table}[t]
\centering
\renewcommand{\arraystretch}{2.0} 
\caption{Overview of popular agent protocols.}
\resizebox{\textwidth}{!}{
\begin{tabular}{c|c|c|c|c|c|c}
\toprule[1.1pt] 
\rowcolor{rowshade}
\textbf{Entity} & \textbf{Scenarios} & \textbf{Protocol} & \textbf{Proposer} & \textbf{Application Scenarios} & \textbf{Key Techniques} & \textbf{Development Stage} \\
\hline
\multirow{2}{*}{\makecell{\textbf{Context-}\\\textbf{Oriented}}} & \makecell{\textbf{General-}\\\textbf{Purpose}} & \makecell{MCP\\\cite{anthropic2025}} & Anthropic & Connecting agents and resources & RPC, OAuth & Factual Standard\\
\cline{2-7}
 & \makecell{\textbf{Domain-}\\\textbf{Specific}} & \makecell{agent.json\\\cite{agentsjson2025}}  & Wildcard AI & \makecell{Offering website \\information to agents} & /.well-known & Drafting\\
 \hline
 \multirow{12}{*}{\makecell{\textbf{Inter-}\\\textbf{Agent}}} &  \multirow{6}{*}{\makecell{\textbf{Genreal-}\\\textbf{Purpose}}} & 
\makecell{A2A\\\cite{a2a2025}}  & Google & Inter-agent communication & RPC, OAuth & Landing\\
\cline{3-7}
 &  & \makecell{ANP\\\cite{anp2024}} & ANP Community  & \makecell{Inter-agent communication} & JSON-LD, DID & Landing\\
 \cline{3-7}
 &  & \makecell{AITP\\\cite{aitp}}  & NEAR Foundation & \makecell{Inter-agent communication} & Blockchain, HTTP & Drafting\\
  \cline{3-7}
 &  & \makecell{AComP\\\cite{agentcommunicationptl}}  & IBM & \makecell{Multi agent system communication} & OpenAPI & Drafting\\
   \cline{3-7}
 &  & \makecell{AConP\\\cite{agentconnectptl}} & Langchain & \makecell{Multi agent system communication} & OpenAPI, JSON & Drafting\\
   \cline{3-7}
 &  & \makecell{Coral\\cite{agentcoralprotocol}} & The Coral Community & 
 \makecell{Multi agent system communication} & - & Drafting\\
   \cline{3-7}
 &  &
\makecell{Agora\\\cite{marro2024scalablecommunicationprotocolnetworks}}  & University of Oxford  & \makecell{Meta protocol between agents} & Protocol Document & Concept\\
 \cline{2-7}
 & \multirow{6}{*}{\makecell{\textbf{Domain-}\\\textbf{Specidic}}} & \makecell{LMOS\\\cite{lmos2025}} & Eclipse Foundation & Internet of things and agents & WOT, DID & Landing\\
  \cline{3-7}
 &  & \makecell{Agent Protocol\\\cite{agentprotocol2025}} & AI Engineer Foundation & Controller-agent interaction & RESTful API & Landing\\
    \cline{3-7}
 &  & \makecell{LOKA\\\cite{ranjan2025lokaprotocoldecentralizedframework}}  & CMU & Decentralized agent system & DECP & Concept\\
    \cline{3-7}
 &  & \makecell{PXP\\\cite{srinivasan2024implementationapplicationintelligibilityprotocol}}  & BITS Pilani & Human-agent interaction & - & Concept\\
     \cline{3-7}
 &  & \makecell{CrowdES\\\cite{bae2025continuouslocomotivecrowdbehavior}}  & GIST.KR & Robot-agent interaction & - & Concept\\
      \cline{3-7}
 &  & 
\makecell{SPPs\\\cite{gąsieniec2024anonymousdistributedlocalisationspatial}}  & University of Liverpool & Robot-agent interaction & - & Concept\\
\bottomrule[1.1pt]
\end{tabular}}
\label{tab:overview-of-protocols}
\end{table}

\subsubsection{General-Purpose Protocols}
General-purpose agent protocols are designed to accommodate a wide range of entities through a unified protocol paradigm, thereby facilitating diverse communication scenarios.
\paragraph{MCP~\citep{anthropic2025}}  In the realm of this kind of agent protocols, Model Context Protocol (MCP) stands out as a pioneering and widely recognised protocol, initially proposed by An. Therefore, this section focuses on introducing MCP, delving into its principles and applications.

MCP is a universal and open context-oriented protocol for connecting LLM agents to resources consisting of external data, tools and services in a simpler and more reliable way~\citep{anthropic2025}. The high standardization of MCP effectively addresses the fragmentation arising from various base LLMs and tool providers, greatly enhancing system integration. At the same time, the standardisation of MCP also brings high scalability to tool usage for LLM agents, making it easier for them to integrate a wide range of new tools. In addition, the client-server architecture of MCP decouples tool invocation from LLM responses, reducing the risk of data leakage.

The following discussion will proceed to introduce the fundamental structure and process of the MCP protocol. The utilisation of the MCP protocol for tool usage can be characterised by the presence of four distinct components, namely \textbf{Host}, \textbf{Client}, \textbf{Server} and \textbf{Resource}. 

\begin{itemize}
    \item \textbf{Host} refers to LLM agents, responsible for interacting with users, understanding and reasoning through user queries, selecting tools, and initiating \textbf{strategic context request}. Each host can be connected to multiple clients.
    \item \textbf{Client} is connected to a host and responsible for providing descriptions of available resources. The client also establishes a one-to-one connection with a server and is responsible for initiating \textbf{executive context request}, including requiring data, invoking tools, and so on.
    \item \textbf{Server} is connected to the resource and establishes a one-to-one connection with the client, providing required context from the resource to the client.
    \item \textbf{Resource} refers to data (e.g., local file systems), tools (e.g., Git), or services (e.g., search engines) provided locally or remotely.
\end{itemize}

In the \textit{initial phase} of a complete MCP invocation cycle, when faced with a user query, the host employs the LLMs' understanding and reasoning capabilities to infer the context necessary to formulate a response to the query. Concurrently, the multiple clients connected to the host provide natural language descriptions of the available resources. Based on the information available, the host determines which resources to request context from and initiating a strategic context request to the corresponding client. In the \textit{request phase} of the MCP invocation cycle, the client sends an executive context request to the corresponding server, encompassing operations such as data modifications or tool invocations. Upon receiving the client's request, the server operates on the resources as specified and subsequently transmits the obtained context to the client, which then passes it on to the host. In the \textit{response phase} of the MCP cycle, the host combines the context obtained to formulate a reply to the user query, thereby completing the cycle.

MCP addresses fragmentation in the LLM ecosystem by introducing a publicly standardized invocation protocol that decouples tool usage from specific base LLM providers and context providers interfaces. By aligning tool invocation with MCP, base LLM providers avoid implementing proprietary formats, thereby enabling greater interoperability and seamless switching between models. Concurrently, context providers support MCP through a one-time integration process, thereby enabling any MCP-compatible LLM agent to access their services. This standardization has been shown to have a significant impact on development and maintenance costs, whilst concomitantly improving scalability and cross-platform compatibility.

Additionally, MCP reduces data security risks arised from the coupling of tool invocations in function-calling style with LLM responses.

Specifically, when requesting context, LLMs generate a complete executable function call, which is then invoked by an external tool. However, in circumstances where the context necessitates private user information for verification (such as account credentials), LLMs may request this information from the user and include it within the generated function call. In this scenario, users of cloud-based LLMs are required to upload their private information to the cloud, posing significant data security risks. Consequently, decoupling tool invocations from LLM responses to mitigate these security concerns is one of the challenges currently faced by LLM agents.

MCP enhances privacy and security in context acquisition by decoupling tool invocation from LLM responses. Instead of executing function calls directly, which may contain sensitive user data, the LLM specifies required resources and parameters, which are then handled by the local client. The client is responsible for constructing and executing the actual context request, and for managing any necessary user authorisation on a local level. Consequently, the confidentiality of sensitive information can be maintained by storing it offline, thereby mitigating the risk of data leakage. This architecture empowers users to exercise control over the contextual data shared with the LLM, thereby mitigating privacy concerns while ensuring the continued efficacy of the tool.

MCP represents a significant step toward standardizing interaction between LLM agents and external resources. By providing a unified protocol for context acquisition and tool invocation, MCP reduces fragmentation across both base LLM providers and resource interfaces. Its client-server architecture enhances interoperability, scalability, and privacy, making it a foundational framework for building robust and secure LLM agent systems.

\subsubsection{Domain-Specific Protocols}
In addition to general-purpose agent protocols, some protocols focus on specific domains to enable targeted enhancements within those areas.

\paragraph{agents.json~\citep{agentsjson2025}} The agents.json  specification is an \textbf{open-source}, \textbf{machine-readable contract format} designed to bridge the gap between traditional \textbf{APIs} and \textbf{AI agents}. Built atop the OpenAPI standard, it enables websites to declare AI-compatible interfaces, authentication schemes, and multi-step workflows in a structured JSON file, typically hosted at \texttt{/.well-known/agents.json}. Unlike conventional OpenAPI specs tailored for human developers, agents.json introduces constructs such as \textit{flows}—predefined sequences of API calls—and \textit{links} that map data dependencies between actions, facilitating reliable orchestration by large language models (LLMs). The design emphasizes statelessness, minimal modifications to existing APIs, and optimization for LLM consumption. By providing a clear, standardized schema for agent interaction, agents.json simplifies integration, reduces the need for prompt engineering, and enhances the discoverability and usability of APIs in agentic contexts.

\subsection{Inter-Agent Protocols}
With the development of large language models (LLMs) and agent technologies, increasing attention has been directed toward overcoming the limitations of single-agent capabilities to address more complex tasks.  Interest in multi-agent collaboration has surged significantly. In some large-scale, complex and inherently decomposable or distributed tasks, multi-agent methods can improve efficiency, reduce costs, and offer better fault tolerance and flexibility, often outperforming single-agent systems in overall performance ~\citep{stone2000multiagent, dorri2018multi}. Interaction among agents is a crucial component in Multi-Agent System (MAS). However, most current MAS frameworks directly embed agents into the system structure without a clearly defined standard for agent interaction methods, which will hinder the development of multi-agent systems. Therefore, there is a growing need to establish a standardized protocol governing the interaction among agents, referred to as the Inter-Agent Protocol.

The protocol should effectively address issues such as agents discovery, information sharing, and the standardization of communication methods and interfaces, thereby providing a unified protocol for inter-agent interaction. In practical application, agents deployed across different platforms and belonging to different vendors often have different skills and capabilities, and may need to interoperate to fulfill users' specific requests. Various types of communication such as discussion, negotiation, debate, and collaboration may occur, all of which involve the exchange of information among two or more agents. The Inter-Agent Protocol plays a pivotal role in enabling and managing these interaction scenarios.


Analogous to context-oriented agent protocols, inter-agent protocols can also be classified into general-purpose and domain-specific categories based on their application scenarios.

\subsubsection{General-Purpose Protocols}

Several inter-agent protocols have already been proposed, including the Agent Network Protocol (ANP)~\citep{anp2024}, Google's Agent2Agent Protocol (A2A)~\citep{a2a2025}, the Agent Interaction \& Transaction Protocol~\citep{aitp}, the Agent Connect Protocol (AConP)~\citep{agentconnectptl}, and the Agent Communication Protocol (AComP)~\citep{agentcommunicationptl}. Although all of them construct protocols focused on the interaction of agents, they vary in their problem domains, application scenarios, and implementation strategies. The following parts will discuss protocols above.

\definecolor{rowshade}{RGB}{240,240,240} 
\begin{table}[t]
\centering
\renewcommand{\arraystretch}{1.4}
\caption{Comparison of different general-purpose inter-agent protocols. Development Stages assessed in Apr. 2025.}
\resizebox{\textwidth}{!}{
\begin{tabular}{P{4.2cm}|P{5.0cm}|P{4.5cm}|P{3.5cm}|P{3.5cm}}
\toprule[1.1pt]
\rowcolor{rowshade}
\textbf{Inter-Agent Protocol} & \textbf{Core Problem} & \textbf{Application Scenarios} & \textbf{Key Techniques} & \textbf{Development Stage} \\
\midrule
\makecell[l]{\textcolor{blue}{ANP}\\(Agent Network Protocol)} & 
Cross-Domain Agent Communication & 
Agent on the Internet & 
JSON-LD, DID & 
Landing \\
\midrule
\makecell[l]{\textcolor{blue}{A2A}\\(Agent2Agent Protocol)} & 
Complex Problem Solving of Agents & 
\makecell[l]{Inter Agent\\Collaboration} & 
RPC, OAuth & 
Landing \\
\midrule
\makecell[l]{\textcolor{blue}{AITP}\\(Agent Interaction \& \\Transaction Protocol)} & 
\makecell[l]{Agent Communication and\\Value Exchange} & 
\makecell[l]{Agents Secure Transactions\\and Interactions} & 
Blockchain, HTTP & 
Drafting \\
\midrule
\makecell[l]{\textcolor{blue}{AConP}\\(Agent Connect Protocol)} & 
\makecell[l]{Standardize Interface to Invoke\\and Configure Agents} & \makecell[l]{Agent Communication and\\Value Exchange} & 
OpenAPI, JSON & 
Drafting \\
\midrule
\makecell[l]{\textcolor{blue}{AComP}\\(Agent Communication \\Protocol)} & 
\makecell[l]{Standardize Practical, Valuable\\Communication Features} &\makecell[l]{Agent Communication and\\Value Exchange} & 
OpenAPI& 
Drafting \\
\midrule
\makecell[l]{\textcolor{blue}{Coral}} & 
\makecell[l]{Decentralized Collaborative \\Infrastructure for AI Agents} & The Internet of Agents & 
- & 
Drafting \\
\midrule
\makecell[l]{\textcolor{blue}{Agora}} & 
\makecell[l]{Meta Protocol\\Communication Flexibility} & 
\makecell[l]{Inter Agent\\Collaboration} & Oxford & Drafting \\
\bottomrule[1.1pt]
\end{tabular}}
\label{tab:interaction-comparison-2}
\end{table}

\paragraph{Agent Network Protocol~\citep{anp2024}} Agent Network Protocol (ANP)  is an open-source agent protocol developed by the open-source technology community, aiming to enable interoperability among various agents across heterogeneous domains. Its vision is to define standardized connection mechanisms between agents and to build an open, secure, and efficient collaborative network for billions of agents. Just as human interaction leading to the emergence of the Internet, the consensus within the agent network is similarly inspired. However, realizing such a network requires designing agent-specific infrastructures tailored to the unique communication and coordination needs of agents. The core principles of ANP are:
\begin{itemize}
  \item \textbf{Interconnectivity}: Enable communication between all agent, break down data silos and ensure AI has access to complete contextual information
  \item \textbf{Native Interfaces}: Agents are not constrained by human interaction habits such as screen capturing or manual clicking when accessing the Internet. Instead, they should interact with the digital world through APIs and protocols and optimize for machine-to-machine communication.
  \item \textbf{Efficient Collaboration}: Agents can establish a more cost-effective and efficient collaboration network by leveraging automatic-organization and automatic-negotiation mechanisms.
\end{itemize}

ANP consists of three core layers:
\begin{itemize}
    \item \textbf{Identity and Encrypted Communication Layer}: This layer leverages the \textbf{W3C DID (Decentralized Identifiers)} standard to establish a decentralized identity authentication mechanism, enabling trustless, end-to-end encrypted communication. This ensures that agents across different platforms can authenticate each other securely.
    \item \textbf{Meta-Protocol Layer}: Serving as a \textit{protocol of protocols}, this layer enables agents to autonomously negotiate and coordinate communication protocols using natural language, such as Agora~\citep{marro2024scalablecommunicationprotocolnetworks}. It supports the dynamic adaptation of communication protocols to accommodate varying interaction needs.
    \item \textbf{Application Protocol Layer}: This layer is responsible for defining standardized protocols that regulate the discovery of agents by other agents on the Internet, the description of the information, capabilities and interfaces offered by these agents, and the application protocols used to accomplish domain-specific tasks.
\end{itemize}

The workflow can be simply described as follows. A local agent first retrieves a list of other agents via a standardized discovery path. Then it accesses the agent description files referenced in the list. Based on the information provided in the description file, the agent initiates interaction by utilizing the required interfaces, constructing properly formatted requests, appending authentication credentials, sending the requests, and finally processes the corresponding responses.

In terms of the significance of ANP, it introduces an innovative solution for agent network communication, marking the creative concept of the \textit{Internet of Agents}. Future directions include optimizing cross-platform identity authentication to improve scalability and practicality, exploring more suitable agent communication protocols to improve data exchange efficiency and reliability, and investigating the potential application of blockchain technology in agent networks, particularly in the areas of decentralized identity management and economic incentive mechanisms.

\paragraph{Agent2Agent Protocol~\citep{a2a2025}}
Agent2Agent (A2A) Protocol is a kind of agent collaboration protocol proposed by Google, designed to enable seamless agent collaboration regardless of underlying frameworks and vendor implementations. It simplifies the integration of agents within different environments and provides core functions required to buildecure, like enterprise-grade agent ecosystem. These functions include capability discovery, user experience negotiation, task and state management, and secure collaboration. Therefore, A2A is specifically designed to support complex inter agent collaboration. The key principles of ANP are:
\begin{itemize}
    \item \textbf{Simplicity}: A2A emphasizes reusing existing standards. For example, it adopts HTTP(S) as the transport layer, JSON-RPC 2.0 as the messaging format, and Server-Sent Events (SSE) for streaming. This lightweight protocol design reduces both the learning curve and implementation complexity.
    \item \textbf{Enterprise Readiness}: The protocol is designed with built-in considerations for authentication, authorization, security, privacy, traceability, and observability. Agents can be treated as enterprise-grade applications, ensuring robustness and security in production environments. 
    \item \textbf{Async-First Architecture}: A2A is centered around the concept of Task, and supports long-running asynchronous workflows, including scenarios involving multi-turn human-in-the-loop interactions. It supports various asynchronous patterns such as polling, SSE-based updates, and push notifications, enabling real-time feedback, notifications and task status updates.
    \item \textbf{Modality Agnostic}: A2A natively supports text, files, forms, media formats such as audio/video streams and embedded frames (iframes). This reflects the multi-modal nature of agent environments.
    \item \textbf{Opaque Execution}: Agent interactions in A2A do not required to share thoughts, plans, or tools. The focus remains on context, state, instructions, and data, preserving implementation privacy and intellectual property. However, task-related metadata is shared, resulting in a semi-transparent collaboration with potential risks of resource exposure.
\end{itemize}

The key concepts defined in the A2A protocol include Agent Card, Task, Artifact, Message, and Parts, which together structure the description of agents and collaborative workflows.

A2A facilitates communication between a \textit{client} agent and a \textit{remote} agent. A client agent is responsible for formulating and communicating tasks, while the remote agent is responsible for acting on those tasks in an attempt to provide the correct information or take the correct action. The workflow can be described as follows. First, remote agents advertise their capabilities using an “Agent Card” in JSON format, allowing the client agent to identify the best agent that can perform the task. Then they leverage A2A to communicate with each other to complete the task. The task object can be completed immediately or run for a long time. Finally, the output of the task is responsed by remote agent in artifact.

A2A advances agent interoperability by introducing a standardized protocol for agent communication. It has demonstrated initial success in enabling seamless agent collaboration within enterprise environments, laying the foundation for broader and more comprehensive agent cooperation. This progress provides both a technical pathway and conceptual framework for the future development of interoperable multi-agent systems.

\paragraph{Agent Interaction \& Transaction Protocol (AITP)~\citep{aitp}} 
The AITP enables AI agents to communicate securely across trust boundaries, while providing extensible mechanisms for structured interactions. It supports autonomous, secure communication, negotiation, and value exchange between agents belonging to different organizations or individuals. For example, in a flight booking scenario, a personal assistant agent can use AITP to directly interact with airline booking agents to exchange flight, passenger, and payment information, instead of navigating airline websites. In AITP, agents communicate through \textbf{Threads}, which are transmitted over a \textbf{Transport} layer, and exchange structured data via \textbf{Capabilities} tailored to specific operations. What distinguishes AITP is its explicit focus on \textbf{enabling agent interactions across trust boundaries}, addressing challenges of identity, security, and data integrity with \textbf{Blockchain} in decentralized multi-agent environments.

\paragraph{Agent Connect Protocol (AConP)~\citep{agentconnectptl}} 
The Agent Connect Protocol (AconP) defines a standard interface for invoking and configuring agents. It provides a set of callable APIs covering five key aspects: \textbf{agent retrieval, execution (run), interruption and resumption, thread management (thread run), and output streaming}. Together, these APIs constitute the usage flow for interacting with agent. The necessary agent information for invoking an agent is stored in the \textbf{Agent ACP Descriptor}, which uniquely identifies an agent, describes its capabilities, and specifies how these capabilities can be consumed. Strictly speaking, AconP defines a standard interface for connecting to and utilizing agent, rather than explicitly facilitating inter-agent interaction. However, by leveraging the ACP Descriptor in combination with the API set, agents can also be interconnected and collaborate with one another through AconP.

\paragraph{Agent Communication Protocol (AComP)~\citep{agentcommunicationptl}} 
It is a universal protocol which \textbf{allow agents to communicate and collaborate across different teams, frameworks, technologies and organizations}, transforming the fragments landscape of today's AI Agents into inter-connected team mates. 
Rather than imposing strict specifications immediately, AComP emphasizes practical, useful features first, and will standardize features that demonstrate value, ensuring broader adoption and long-term compatibility. It doesn’t manage workflows, deployments, or coordination between agents. Instead, AcomP enables orchestration across diverse agents by standardizing how they communicate.
The currently lastest version holds four key features: REST-based communication (use HTTP patterns), no SDK required (but available), offline discovery and async-first, sync supported, enabling RESTful, streaming-compatible architecture, structured multipart messages (MIME) and token-based security. 

\paragraph{The Coral Protocol~\citep{agentcoralprotocol}}
The Coral Protocol is an \textbf{open and decentralized collaborative infrastructure designed to address the interoperability challenges among proprietary AI agents deployed by different organizations}. It provides a \textbf{unified framework for communication, coordination, trust, and payments among AI agents}, enabling efficient collaboration across diverse domains and vendors. Key features of the protocol include a threaded communication model, the use of the Coral Server as a coordination layer, and MCP as a standardized agent communication protocol. Additionally, agents are defined by a structured identity, an assigned role, and scoped memory, which establish how agents interact with the system and with one another, enabling trust, modularity, and controlled information sharing. The protocol supports secure and efficient collaboration for agents built on any framework, positioning itself as a potentially practical protocol for integrating agents into software systems in the future.

\paragraph{Agora~\citep{marro2024scalablecommunicationprotocolnetworks}} Before the advent of LLM agents, researchers in the field of computer science devoted decades exploring the design of communication paradigms for agents~\citep{gilbert2019agent}. The emergence of LLM agents has reinvigorated the discourse surrounding agent communication protocols. LLMs have shown remarkable improvements in both following instructions in natural language~\citep{wei2022chain} and handling structured data~\citep{collins2022structuredflexiblerobustbenchmarking}. Concurrently, LLMs have exhibited remarkable proficiency in a variety of real-world tasks~\citep{pyatkin2022reinforced,zhong2023study,wei2022chain}. According to~\cite{hu2022lora} and \cite{marro2024scalablecommunicationprotocolnetworks}, specialized LLMs demonstrate superior performance in comparison to general-purpose LLMs, underscoring the considerable potential of agent networks based on heterogeneous LLMs. The distinguishing characteristics of heterogeneous LLMs primarily encompass architecture, capabilities and usage policies. However, agent networks based on heterogeneous LLMs face an \textbf{Agent Communication Trilemma}, struggling to balance \textbf{versatility}, \textbf{efficiency}, and \textbf{portability}~\citep{marro2024scalablecommunicationprotocolnetworks}.

\begin{itemize}
    \item \textbf{Versatility}: Communication between agents needs to support various types and formats of messages to ensure the versatility to support a broad spectrum of tasks and scenarios.
    \item \textbf{Efficiency}: The computational cost of employing agents and facilitating communication should be maintained at a minimum to ensure the system operates efficiently.
    \item \textbf{Portability}: The implementation of the communication protocol should require minimal effort from human programmers, facilitating greater participation in the communication network by enormous agents.
\end{itemize}

In LLM agent communication, versatility, efficiency, and portability form the Agent Communication Trilemma. Versatility requires communication to support various message types and formats to accommodate different task scenarios, yet this increases the complexity of the protocol, raises implementation difficulty and costs, and results in diminished portability. Efficiency demands low computational and network costs for communication and minimize ambiguity that could lead to potential errors, but highly flexible communication mechanisms frequently entail a substantial computational overhead, such as frequent use of natural language communication. Portability requires the protocol to be easy to implement and deploy, but complex and flexible protocols demand significant programming efforts, consequently making the application across different agents difficult and time-consuming. The three factors are interdependent, making it challenging to optimize all of them simultaneously.

In order to address the Agent Communication Trilemma, Agora leverages the capabilities of LLMs in natural language understanding, code generation, and autonomous negotiation, thereby enabling agents to adopt various communication protocols based on context. Frequent communications employ structured protocols to ensure efficiency, while infrequent ones rely on structured data with routines generated by the LLM. In the event of rare communications or failures, LLM agents transition to natural language, a change that can also facilitate protocol negotiation. Agora introduces Protocol Documents (PDs), which are plain-text protocol descriptions that allow agents to autonomously negotiate, implement, adapt, and even create new protocols without human intervention.

Agora has been designed to adapt to various scenarios by supporting multiple communication methods, including traditional structured protocols, LLM routines, and natural language communication, thus meeting the versatility. For frequent communication tasks, it prioritizes efficient traditional protocols and LLM routines to minimise computation and latency, using natural language only when necessary, thus balancing versatility and efficiency. Its design enables agents to autonomously negotiate, implement, and use protocols, thereby reducing dependence on human programming. PDs facilitate protocol sharing and provide support for various scenarios and LLMs, enhancing compatibility and scalability, effectively addressing the Agent Communication Trilemma.

\subsubsection{Domain-Specific Protocols}

Domain-specific protocols serve as \textbf{tailored communication and coordination mechanisms} that govern interactions between intelligent agents and their counterparts across distinct operational domains. These protocols are designed to address the unique requirements and constraints of each interaction context, ensuring robust, interpretable, and ethically aligned behavior. In this section, we categorize domain-specific protocols into three primary branches: (1) \textit{Human–Agent Interaction Protocols}, which focus on fostering mutual intelligibility and trust; (2) \textit{Robot–Agent Interaction Protocols}, which emphasize spatial reasoning and behavioral coordination in physical environments; and (3) \textit{System–Agent Interaction Protocols}, which facilitate scalable, interoperable, and secure multi-agent ecosystems. Each category encapsulates specialized protocol frameworks that address the nuances of communication, identity, decision-making, and task execution in their respective domains.

\vspace{1ex}
\noindent\textbf{3.2.2.1 Human–Agent Interaction Protocol}\label{Sec:Human-Agent Interaction}
\vspace{1ex}

Human–Agent Interaction Protocols are specifically designed to enable meaningful, transparent, and context-aware communication between human users and intelligent agents. In domains where interpretability, collaboration, and ethical decision-making are critical, such protocols provide the necessary structure for aligning machine behavior with human intentions and expectations. This category emphasizes both the cognitive alignment (i.e., intelligibility of predictions and reasoning processes) and the normative alignment (i.e., ethical and accountable interactions) between humans and agents. The following protocols illustrate two complementary approaches to this goal: the PXP protocol focuses on mutual intelligibility in task-oriented dialogues, while the LOKA protocol establishes a decentralized foundation for identity, trust, and ethical coordination in heterogeneous multi-agent systems.

\paragraph{PXP Protocol~\citep{srinivasan2024implementationapplicationintelligibilityprotocol}} The PXP protocol (\textbf{P}redict and e\textbf{X}plain \textbf{P}rotocol), a cornerstone of domain-specific human-agent interaction protocols, is designed to facilitate \textbf{bidirectional intelligible interactions between human experts and machine agents powered by LLMs}. This protocol employs a finite-state machine model, enabling agents to communicate through messages tagged with four labels: \textit{RATIFY}, \textit{REFUTE}, \textit{REVISE}, and \textit{REJECT}. These tags are determined based on the agreement or disagreement of predictions and explanations exchanged between the agents. The implementation of the PXP protocol involves a blackboard system and a scheduler that alternates between human and machine agents. The protocol has been experimentally validated in two distinct domains: \textbf{radiology diagnosis} and \textbf{drug synthesis pathway planning}. These experiments demonstrated the protocol's capability to capture one-way and two-way intelligibility in human-LLM interactions, providing empirical support for its potential in designing effective human-LLM collaborative systems.

\paragraph{LOKA Protocol~\citep{ranjan2025lokaprotocoldecentralizedframework}} The LOKA (\textbf{L}ayered \textbf{O}rchestration for \textbf{K}nowledgeful \textbf{A}gents) Protocol introduces a comprehensive decentralized framework designed to address the challenges of \textbf{identity, accountability, and ethical alignment in AI agent ecosystems}. It proposes a \textbf{Universal Agent Identity Layer (UAIL)} to assign unique, verifiable identities to AI agents, facilitating secure authentication, accountability, and interoperability. Building on this foundation, the protocol incorporates intent-centric communication protocols to enable semantic coordination across diverse agents. A key feature is the \textbf{Decentralized Ethical Consensus Protocol (DECP)}, which allows agents to make context-aware decisions grounded in shared ethical baselines. Anchored in emerging standards such as \textbf{Decentralized Identifiers (DIDs)}, \textbf{Verifiable Credentials (VCs)}, and post-quantum cryptography, LOKA aims to provide a scalable and future-resilient blueprint for multi-agent AI governance. By embedding identity, trust, and ethics into the protocol layer itself, LOKA establishes a foundation for responsible, transparent, and autonomous AI ecosystems operating across digital and physical domains.

\paragraph{Web-Agent Protocol~\citep{ota_webagent_2025}} The \textbf{Web-Agent Protocol (WAP)} defines a standardized interaction framework for coordinating user–agent–browser interactions through \textbf{record-and-replay mechanisms}. Designed to enhance the automation and auditability of web-based tasks, WAP decouples the recording of user interactions from their execution, enabling efficient reuse and precise task reproduction. The protocol operates in four stages: (1) user interactions are captured using the \textit{OTA-WAP Chrome Extension}; (2) a backend service processes the raw event stream into structured action lists, supporting both exact and intelligent replay; (3) the action sequences can be embedded within contextual environments using the \textit{MCP (Message Context Protocol)} server; and (4) these actions are executed in the browser via the \textit{WAP-Replay} component. Key use cases include web automation, agent-based UI testing, and behavioral modeling for intelligent assistants. By providing a modular and interpretable way to integrate human inputs into web agents’ workflows, WAP complements existing intelligibility protocols and contributes to the development of responsive and accountable web-based agent systems.

\vspace{1ex}
\noindent\textbf{3.2.2.2 Robot-Agent Interaction Protocol}\label{Sec:Robot-Agent Interaction}
\vspace{1ex}

Robot–Agent Interaction Protocols address the challenges of coordination, perception, and spatial reasoning in physical environments where intelligent agents—particularly embodied robots—must interact with one another and with dynamic surroundings. These protocols are essential for enabling distributed decision-making, real-time environmental adaptation, and safe navigation in complex multi-agent systems. They must account for uncertainties in sensor input, partial observability, and limited communication bandwidth while still supporting robust group behaviors. In this section, we present two representative approaches: the CrowdES protocol, which focuses on simulating and adapting to realistic crowd dynamics in robot-populated environments, and the Spatial Population Protocols, which offer distributed solutions for achieving geometric consensus among anonymous robotic agents in decentralized systems.

\paragraph{CrowdES~\citep{bae2025continuouslocomotivecrowdbehavior}} The CrowdES framework introduces a novel interaction protocol designed for \textbf{continuous and realistic crowd behavior generation}, particularly relevant for robot-agent interactions. This protocol integrates a \textbf{crowd emitter} and a \textbf{crowd simulator} to dynamically populate environments and simulate diverse locomotion patterns. The crowd emitter uses diffusion models to assign individual attributes, such as agent types and movement speeds, based on spatial layouts extracted from input images. The crowd simulator then generates detailed trajectories, incorporating intermediate behaviors like collision avoidance and group interactions using a Markov chain-based state-switching mechanism. This protocol allows for real-time control and customization of crowd behaviors, enabling robots to navigate and interact within dynamic, heterogeneous environments. The implementation leverages advanced techniques like diffusion models for agent placement and switching dynamical systems for behavior augmentation, ensuring both realism and flexibility in robot-agent interactions.

\paragraph{Spatial Population Protocols~\citep{gąsieniec2024anonymousdistributedlocalisationspatial}} Spatial Population Protocols (SPPs) are proposed for solving the \textbf{distributed localization problem (DLP) among anonymous robots}. This protocol enables robots to \textbf{reach a consensus on a unified coordinate system through pairwise interactions}, even when they start in arbitrary positions and coordinate systems. The key innovation lies in the ability of each robot to memorize one or a fixed number of coordinates and to query either the distance or the vector between itself and another robot during interactions. The protocol is implemented in three variants:
\begin{itemize}
    \item \textbf{Self-stabilising Distance Query Protocol:} This protocol adjusts labels based on pairwise distances, achieving $\epsilon$-stability in $O(n)$ parallel time. It is particularly effective in random configurations but faces challenges in certain hard instances.
    \item \textbf{Leader-based Distance Query Protocol:} Utilizing a leader to anchor the coordinate system, this protocol stabilizes in sublinear time $O(n)$ through a multi-contact epidemic process, significantly improving efficiency.
    \item \textbf{Self-stabilising Vector Query Protocol:} This variant leverages vector queries to achieve superfast stabilization in $O(\log n)$ parallel time, demonstrating the power of richer geometric information in interactions.
\end{itemize}
These protocols provide a robust framework for robot-agent interactions, enabling efficient and accurate localization in distributed systems.

\vspace{1ex}
\noindent\textbf{3.2.2.3 System-Agent Interaction Protocol}\label{Sec:System-Agent Interaction}
\vspace{1ex}

System–Agent Interaction Protocols provide the foundational infrastructure for orchestrating, managing, and integrating AI agents within complex digital ecosystems. This category encompasses protocols that address the challenges of agent discovery, interoperability, lifecycle management, and secure communication. Notably, the \textbf{Language Model Operating System (LMOS)} offers a comprehensive framework for building and operating multi-agent systems, emphasizing openness and scalability. The agents.json specification introduces a standardized, machine-readable format for declaring AI-compatible interfaces and workflows, facilitating seamless integration between traditional APIs and AI agents. Meanwhile, the \textbf{Agent Protocol} defines a framework-agnostic communication standard, enabling control consoles to manage agent operations effectively. Together, these protocols establish a robust foundation for the development and deployment of interoperable, scalable, and secure AI agent ecosystems.

\paragraph{LMOS~\citep{lmos2025}} The \textbf{Language Model Operating System (LMOS)}  protocol, developed under the Eclipse Foundation, provides a foundational architecture for building an \emph{Internet of Agents (IoA)}—a decentralized, interoperable, and scalable ecosystem where AI agents and tools can be published, discovered, and interconnected regardless of their underlying technologies. Inspired by open protocols like Matter/Thread and ActivityPub, LMOS is structured into three layers: (1) the \textbf{Application Protocol Layer}, which standardizes agent discovery and interaction using JSON-LD and semantic models; (2) the \textbf{Transport Protocol Layer}, which enables context-aware negotiation of communication protocols (e.g., HTTP, MQTT, AMQP); and (3) the \textbf{Identity and Security Layer}, which ensures secure, verifiable identities via W3C DIDs and supports schemes like OAuth2. Key components include decentralized agent/tool descriptions, metadata propagation mechanisms, group management protocols, and flexible agent communication interfaces. LMOS is implemented as an open-source, cloud-native platform integrated with tools such as ARC, LangChain, and LlamaIndex. Use cases span domains such as customer service and manufacturing, where agents autonomously coordinate across tools and organizations to resolve issues and optimize operations.

\paragraph{Agent Protocol~\citep{agentprotocol2025}} The \textbf{Agent Protocol} is an \textbf{open-source}, \textbf{framework-agnostic communication standard} designed to enable seamless interaction between \textbf{control consoles} and \textbf{AI agents}. Built on OpenAPI v3, it defines a unified interface for executing key agent lifecycle operations—starting, stopping, and monitoring agents. The protocol introduces core abstractions such as \textit{Runs} for task execution, \textit{Threads} for managing multi-turn interactions, and \textit{Store} for persistent, long-term memory. By standardizing these functionalities, Agent Protocol empowers developers to orchestrate heterogeneous agents across diverse systems, promoting interoperability, scalability, and operational transparency in multi-agent environments.

There is some relationship between Inter-Agent Protocol and Context-Oriented Protocol. Within context-oriented interactions, interactive tools can be regarded as low-autonomy agents. Conversely, in agent-to-agent interactions, the communicating agents can also be viewed as tools with higher autonomy, designed to accomplish specific intelligent tasks. Unlike traditional tools connected through protocols such as MCP, an agent acting as a tool is also capable of being a task initiator. The linked agent can subsequently issue requests and interact with other agents or traditional tools. At this level of abstraction, a tool essentially represents a specific skill or capability possessed by an agent. In the long term, these two paradigms, the context-oriented interaction and the autonomous agent interaction, may gradually converge and become increasingly homogeneous in their design and application.

\section{Protocol Evaluation and Comparison}

In the rapidly evolving landscape of agent communication protocols, static performance or functionality comparisons quickly become outdated due to the fast-paced iterations in this domain. For instance, MCP introduced in November 2024 initially lacked support for HTTP and authentication mechanisms. By early 2025, it incorporated HTTP Server-Sent Events (SSE) and authentication, and has since transitioned to HTTP Streaming. This evolution mirrors the progression from TCP/IP to HTTP in the internet era, highlighting continuous enhancements in functionality, performance, and security.

Consequently, this section focuses on identifying the critical dimensions and challenges to consider when designing and evaluating LLM agent communication protocols, rather than proposing a specific evaluation benchmark. Drawing inspiration from the seven core metrics observed in the evolution of internet protocols—interoperability, performance efficiency, reliability, scalability, security, evolvability, and simplicity—we examine their applicability to LLM agent protocols. 
As shown in Table~\ref{tab:protocol evalution}, by delineating these evaluative dimensions, this section aims to provide a comprehensive understanding of the considerations essential for the effective design and assessment of LLM agent protocols, thereby contributing to the advancement of intelligent agent systems.

\definecolor{rowshade}{RGB}{240,240,240} 
\begin{table}[t]
\centering
\renewcommand{\arraystretch}{1.4}
\caption{Overview of protocol evaluation from different dimensions.}
\resizebox{\textwidth}{!}{
\begin{tabular}{P{2.7cm}|m{7.0cm}|P{5.2cm}}
\toprule[1.1pt]
\rowcolor{rowshade}
\textbf{Dimension} & \textbf{Description} & \textbf{Key Metric} \\
\midrule
\textbf{Efficiency} & Fast and resource-efficient communication. & \makecell[l]{$\bullet$ Latency \\ $\bullet$ Throughput \\ $\bullet$ Resource Utilization} \\
\midrule
\textbf{Scalability} & 
Stable performance with increasing complexity of tools/agents/networks. & \makecell[l]{$\bullet$ Node Scalability \\ $\bullet$ Link Scalability \\ $\bullet$ Capability Negotiation} \\
\midrule
\textbf{Security} & Trusted interactions via authentication, access control, and data safeguarding. & \makecell[l]{$\bullet$ Authentication Mode Diversity \\ $\bullet$ Role/ACL Granularity \\ $\bullet$ Context Desensitization \\ \;\; Mechanism} \\
\midrule
\textbf{Reliability} & Consistent, accurate, and fault-tolerant communication. & \makecell[l]{$\bullet$ Packet Retransmission \\ $\bullet$ Flow and Congestion Control \\ $\bullet$ Persistent Connections} \\
\midrule
\textbf{Extensibility} & Evolution for new features without disrupting existing systems or compatibility. & \makecell[l]{$\bullet$ Backward Compatibility \\ $\bullet$ Flexibility \& Adaptability \\ $\bullet$ Customization \& Extension} \\
\midrule
\textbf{Operability} & The ease of implementing, managing, and integrating the protocol in real-world systems. & \makecell[l]{$\bullet$ Protocol Stack Code Volume \\ $\bullet$ Deployment \& Configureation \\ \;\; Complexity \\ $\bullet$ Observability} \\
\midrule
\textbf{Interoperability} & Seamless communication and collaboration across diverse platforms, systems, and network environments. & \makecell[l]{$\bullet$ Cross-System \& Cross-Browser \\ \;\; Compatibility \\ $\bullet$ Cross-Network \& Cross-Platform \\ \;\; Adaptability} \\
\bottomrule[1.1pt]
\end{tabular}}
\label{tab:protocol evalution}
\end{table}

\subsection{Efficiency}
Efficiency is a critical dimension for evaluating Agent Protocols, encapsulating their efficiency in managing throughput, minimizing latency, optimizing handshake overhead, and reducing message header size in dynamic, multi-agent and agent-to-tool interactions. In the Agent era, efficiency extends beyond traditional internet protocol metrics to address unique demands like semantic processing, dynamic task coordination \citep{liu2022multi}, and token consumption costs. An ideal protocol should ensure low-latency communication, rapid task completion, and minimal resource overhead, while adapting to the complexity of multi-agent systems.

\paragraph{Latency} Key metrics for assessing efficiency performance include communication latency, measured as the time for a message to be sent, received, and parsed. In Agent Protocols, latency is impacted not only by network transmission \citep{jiang2018lowlatencynetworkinglatencylurks}, but also by semantic processing and protocol-specific overheads. Compared to traditional internet protocols like HTTP, which focus solely on data transfer, Agent Protocols must handle these additional layers. Testing involves measuring round-trip times across network conditions (e.g., low bandwidth, high latency). 

\paragraph{Throughput} Throughput, quantified as the number of messages or tasks processed per second, assesses a protocol’s capacity to handle concurrent interactions in agentic systems. High throughput is essential for scaling to large agent networks, as messages may include complex metadata, which reduces throughput compared to traditional protocols that handle simpler payloads. To evaluate this capability, we provide a metric called \textit{Throughput per Second at concurrency level $N$ (TPS-N)}.
\begin{equation}\label{equ:tps-n}
\text{TPS-N} = \frac{\# \text{Processed Messages}}{\text{Elapsed Time}}
\end{equation}

\paragraph{Resource Utilization} Resource utilization evaluates the protocol’s consumption of computational resources, including header size and token consumption (for LLM-driven tasks), alongside CPU, memory, and bandwidth usage. Token consumption measures the number of tokens consumed by LLM-driven tasks, such as semantic processing or dynamic coordination, unique to Agent-era protocols. Testing involves profiling token usage with LLM monitoring tools across typical tasks (e.g., task assignment and tool query).

\subsection{Scalability}
Scalability refers to an Agent Protocol’s ability to maintain performance and availability as the number of nodes (agents or tools) or connections (links) grows exponentially, ensuring robust operation in increasingly complex and large-scale multi-agent systems. In the Agent era, scalability extends beyond traditional Internet protocol concerns, such as IP address allocation or caching, to include the efficient handling of growing agent populations, dynamic tool integrations, and high-density communication networks. A scalable Agent protocol must support thousands to millions of agents, accommodate diverse workloads, and integrate new functionalities without significant performance degradation.

\paragraph{Node Scalability} Node scalability measures the protocol’s ability to maintain performance as the number of tools, plugins, or agents ($N$) increases, reflecting its capacity to support large-scale networks. While traditional internet protocols, such as IP, utilize CIDR \citep{fuller1993classless} to manage address scalability, Agent Protocols must also handle dynamic node discovery and coordination. Node scalability can be evaluated by analyzing the \textbf{performance degradation curve} (e.g., latency, throughput) as $N$ increases. 

\paragraph{Link Scalability} Link scalability assesses the protocol’s performance as the number of communication links multiplies, critical for dense networks with frequent interactions. This is measured by tracking performance metrics (e.g., throughput, latency) as link density increases, such as in a fully connected mesh of 1,000 agents versus a sparse network. Agent Protocols face challenges due to link-specific overheads, such as task life-cycle management or authentication in each connection, which add computational costs.

\paragraph{Capability Negotiation} Capability negotiation assesses whether the protocol facilitates dynamic agreement on communication protocols, capabilities, or task assignments between agents or between agents and tools, and how effectively it scales with increasing network size. To capture this, we provide a metric called \textit{Capability Negotiation Score (CNS)}, which is evaluated by measuring the success rate and time required for negotiations as the number of nodes increases.
\begin{equation}
\text{CNS} = \frac{\# \text{Successful Negotiations} / \# \text{Negotiation Attempts}}{\text{Average Negotiation Time}}
\end{equation}

\subsection{Security}
Security is a fundamental dimension for evaluating Agent Protocols, ensuring that agent-to-agent and agent-to-tool interactions are protected through robust identity authentication, encryption, and integrity validation. In the Agent era, security extends beyond traditional internet protocol mechanisms, such as SSL/TLS or OAuth, to address the unique challenges of dynamic, decentralized, and semantic-driven agent ecosystems. A secure Agent Protocol must provide reliable identity verification, safeguard data confidentiality, ensure message integrity, and support fine-grained access control.

\paragraph{Authentication Mode Diversity} Authentication mode diversity evaluates the variety of authentication mechanisms supported by the protocol, enabling flexibility for different use cases and security requirements. This metric can be assessed by counting the number of supported modes and their applicability to agent-to-agent and agent-to-tool scenarios.

\paragraph{Role/ACL Granularity} Role/Access Control List (ACL) granularity measures the protocol’s ability to enforce fine-grained access controls, specifying permissions at varying levels, such as field-level, endpoint-level, or task-level. This metric can be evaluated by analyzing the precision of role definitions and ACL configurations, such as whether an agent can access specific data fields in a tool’s response or particular task endpoints.

\paragraph{Context Desensitization Mechanism} Context desensitization mechanism assesses the protocol’s ability to protect sensitive data by anonymizing or redacting contextual information during agent-to-agent or agent-to-tool interactions, minimizing exposure risks. This metric is evaluated by examining the presence and effectiveness of desensitization techniques, such as data masking, tokenization, or selective data sharing.

\subsection{Reliability}
The reliability of Agent Protocol refers to its ability to ensure stable and accurate communication between agents in multi-agent systems. Similar to the Internet Protocol's emphasis on reliable data transmission, Agent Protocol ensures that messages between agents are delivered accurately, completely, and in a timely manner. It employs mechanisms such as message acknowledgment, retransmission, flow control, and congestion control to address potential issues in agent communication, akin to how the Internet Protocol ensures reliable data transmission over networks. Additionally, Agent Protocol incorporates fault tolerance and recovery mechanisms to maintain system stability even when individual agents or communication links fail, much like the Internet Protocol's ability to adapt to network disruptions and reroute data packets to ensure reliable delivery.

\paragraph{Packet Retransmission} Similar to TCP's retransmission mechanism, Agent protocols can implement packet retransmission based on timers. If a sender agent does not receive an acknowledgment (ACK) from the receiver agent within a specified timeframe after sending a message, it will trigger a retransmission. Additionally, the receiver agent can notify the sender agent of packet loss in ACK messages, prompting the sender to retransmit the lost packets, thereby ensuring the completeness and accuracy of data transmission. This can be evaluated by \textit{Automatic Retry Count (ARC)}, which indicates the number of times Agent Protocol automatically retries message transmission when it detects network issues or delivery failures.
\begin{equation}
    \text{Automatic Retry Count (ARC)}=\#\text{message retransmissions when delivery fails}
\end{equation}

\paragraph{Flow and Congestion Control} Agent protocols integrate flow and congestion control mechanisms akin to TCP. For flow control, the receiver communicates its available receive window size to the sender, which dynamically adjusts transmission rates to prevent buffer overflow and data loss. Concurrently, congestion control employs strategies such as slow start and congestion avoidance. The sender initially probes network capacity with a small congestion window, incrementally increasing it based on feedback. Upon detecting packet loss or increased latency indicative of network congestion, the sender reduces its congestion window to diminish transmission rates. These coordinated mechanisms enable efficient data transfer while maintaining network stability and preventing resource exhaustion. This control ability can be evaluated by \textit{Convergence Time (CT)}, which refers to the time required to reach a stable rate at startup, when the available link capacity changes, or when new flows join the bottleneck link.
\begin{equation}
    \text{Convergence Time (CT)}=\text{clock time to reach a stable state when link changes}
\end{equation}

\paragraph{Persistent Connections} Agent protocols can establish persistent connections between agents, allowing communication channels to remain open for multiple data transmissions. Unlike creating a new connection for each interaction, persistent connections eliminate the overhead of frequent connection setup and teardown, reducing latency and improving transmission efficiency. The stability of connections can be evaluated by \textit{Unexpected Disconnection Rate (UDR)}, which gives the number of unexpected disconnections per unit time and \textit{Message-Loss Rate (MLR)}, which refers to the proportion of messages that fail to reach the recipient agent within a specified timeframe out of the total number of messages sent.
\begin{equation}
    \text{Unexpected Disconnection Rate (UDR)}=\frac{\#\text{unexpected disconnections}}{\text{unit time}}
\end{equation}
\begin{equation}
    \text{Message-Loss Rate (MLR)} = \frac{\#\text{messages failing to reach the recipient}}{\#\text{messages sent}}
\end{equation}

\subsection{Extensibility}
The extensibility of Agent Protocol refers to its ability to flexibly adapt to new requirements and technological developments by adding new features or modifying existing functionalities without disrupting backward compatibility. Similar to how the Internet Protocol evolves through mechanisms like custom headers in HTTP or optional fields in IP packets, Agent Protocol also provides a flexible framework that allows for the introduction of new capabilities while maintaining compatibility with existing systems. This ensures that as the needs of multi-agent systems grow and change, the protocol can be extended to accommodate these advancements, ensuring long-term relevance and effectiveness.

\paragraph{Backward Compatibility} Agent Protocol evolves over time. During these iterations, the protocol retains backward compatibility, ensuring that existing functionalities and applications remain unaffected. Users can seamlessly adopt new versions of the protocol without significant adjustments to their existing systems. The backward compatibility can be reflected by \textit{Upgrade Success Rate (USR)}, which indicates the success rate of old clients in maintaining normal interactions with the server after a major version upgrade of Agent Protocol.
\begin{equation}
    \text{Upgrade Success Rate (USR)}=\frac{\#\text{normal interactions after a major upgrade}}{\#\text{total interactions after a major upgrade}}
\end{equation}

\paragraph{Flexibility and Adaptability} Agent Protocol adopts a flexible design, makes it easy to integrate with existing IT stacks and allows the protocol to adapt to new technological advancements and application scenarios. Developers can extend the protocol based on specific needs by adding new fields or semantics. Modality-agnostic design also enhances the protocol's extensibility. Developers can define new communication modalities based on evolving needs while ensuring compatibility with existing text-based communication modes. The flexibility and adaptability can be evaluated by \textit{Automatic Test Pass Rate (ATPR)}, which involves automatically testing the new features or modifications listed in the Agent Protocol changelog and calculating the pass rate.
\begin{equation}
    \text{Automatic Test Pass Rate (ATPR)}=\frac{\#\text{new features passing the test}}{\#\text{new features}}
\end{equation}

\paragraph{Customization and Extension} Agent Protocol allows developers to add custom fields to meet specific application requirements. Developers can extend these fields to enable other agents to discover and interact with them without affecting existing functionalities. It also supports plugin system support. Agent Protocol provides a standardized plugin system, enabling developers to add new features or capabilities through plugins. These plugins can introduce new fields or semantics while maintaining compatibility with the core protocol. 

\subsection{Operability}
The operability of Agent Protocol refers to the ease and efficiency with which it can be implemented, operated, and maintained. Similar to the Internet Protocol, which emphasizes simplicity and clarity in its design to facilitate widespread adoption and deployment, Agent Protocol also prioritizes ease of implementation and use. Its specification is concise and clear, allowing developers to quickly integrate it into their systems. The protocol is framework-agnostic, supporting multiple programming languages and platforms, which reduces implementation complexity and lowers the technical barriers for developers. Additionally, Agent Protocol provides comprehensive documentation, SDKs, and client libraries, offering clear development guidance and tools to help developers efficiently implement the protocol. Furthermore, its layered architecture and modular design enable developers to implement and maintain different components independently, enhancing flexibility and reducing operational complexity. A coarse evaluation metric for operability is the \textit{Number of Dependency Components (NDC)}, which indicates the number of dependency components required for the Agent Protocol.
\begin{equation}
    \text{Number of Dependency Components (NDC)}=\#\text{dependency components required}
\end{equation}

\paragraph{Protocol Stack Code Volume} Agent Protocol is designed as a lightweight API specification, defining a series of endpoints and pre-defined response models with concise logic and clear semantics. Its code volume is relatively small, making it easy to understand and implement. This allows developers to quickly integrate it into their systems without a steep learning curve. For example, the core components of Agent Protocol include the Runs, Threads, and Store modules, which provide complete lifecycle management, state control, and persistent storage capabilities. Developers can focus on business logic implementation rather than worrying about underlying architectural details.

\paragraph{Deployment and Configuration Complexity} Agent Protocol adopts a framework-agnostic approach, supporting multiple programming languages and platforms. This enables developers to implement the protocol using their preferred languages and frameworks. Additionally, the protocol provides comprehensive documentation, SDKs, and client libraries, offering clear development guidance and tools to simplify deployment and configuration.

\paragraph{Observability} Agent Protocol emphasizes observability, providing monitoring tools to help operations personnel track its performance metrics, such as message throughput, latency, and error rates. For example, the LMOS platform's observability module offers enterprise-level monitoring capabilities, meeting compliance requirements. The protocol also provides debugging tools and interfaces to assist developers in diagnosing and resolving issues during agent communication. This ensures stable operation of Agent Protocol and enhances its operability.

\subsection{Interoperability}
The interoperability of Agent Protocol refers to its ability to enable seamless communication between different systems, frameworks, browsers, and other environments. Similar to how the Internet Protocol establishes standards for network communication to ensure data transmission between diverse devices and systems, Agent Protocol defines standardized communication rules and data formats to allow agents developed on various platforms to interact effectively. It enables agents to discover, communicate, and collaborate with one another regardless of their underlying implementation details, much like how different systems and browsers can seamlessly exchange information over the internet. 
The interoperability can be evaluated by \textit{Schema Compatibility Test Pass Rate (SCTPR)}, which reflects how well agents can communicate effectively without version conflicts or data format issues.
\begin{equation}
    \text{Schema Compatibility Pass Rate (SC-PR)} = \frac{\#\text{successful test cases}}{\#\text{total schema compatibility test cases}}
\end{equation}

\paragraph{Cross-System and Cross-Browser Compatibility} Agent Protocol ensures seamless communication between agents running on different operating systems (e.g., Windows, macOS, Linux) and browsers (e.g., Chrome, Firefox, Safari). It provides standardized APIs and communication interfaces that abstract away underlying platform differences. This allows agents to interact using uniform protocols and data formats regardless of the operating system or browser environment. For instance, an agent developed on a Windows system using Chrome can communicate with another agent on a macOS system using Safari. This cross-system and cross-browser compatibility eliminates the need for agents to adapt to specific platform characteristics, enabling broad interoperability.

\paragraph{Cross-Network and Cross-Platform Adaptability} Agent Protocol supports diverse network environments, including local area networks (LANs), wide area networks (WANs), and the internet. It can adapt to varying network conditions, ensuring stable communication between agents even when network parameters change. Additionally, it supports multiple programming languages and platforms, allowing developers to implement agents using their preferred languages and frameworks. This cross-platform and cross-language capability ensures that agents developed on different technical stacks can communicate and collaborate effectively. For example, an agent developed in Python can interact with another agent developed in Java. This adaptability enables agents to operate in heterogeneous network and platform environments, enhancing their interoperability.

\subsection{Evaluation over Protocol Evolution: Case Studies}

In the process of designing and evaluating agent communication protocols, observing their evolutionary trajectory helps to reveal the pathways through which protocols adapt to new requirements and challenges across functionality, performance, and security. The following analysis explores two typical cases—protocol iteration and protocol system evolution—to illustrate how agent protocols continuously evolve in practice to meet emerging demands.

\paragraph{Iteration of MCP} The transition from MCP v1.0 to v1.2 introduced support for HTTP Streaming and authentication (Auth). This change resulted in the following impacts: 
\begin{itemize}
    \item \textbf{Improved Interoperability:} The addition of HTTP support enabled MCP to integrate with a broader range of external systems and services, enhancing the protocol's compatibility and applicability.
    \item \textbf{Enhanced Security:} The implementation of Token-based authentication mechanisms ensured the security of data transmission and the reliability of identity verification.
    \item \textbf{Performance Impact:} While HTTP Streaming facilitated more efficient data transfer, it also potentially introduced new latency factors, requiring re-evaluation and optimization of stream latency performance.
\end{itemize}
This iteration exemplifies the protocol's balancing act between expanding functionality, optimizing performance, and enhancing security, illustrating the multidimensional trade-offs involved in protocol iteration.

\paragraph{Evolution from MCP to ANP and A2A} The progression from MCP to ANP and, ultimately, to A2A represents a shift from a singular functional protocol to a more complex, multi-layered, and multidimensional collaborative architecture:
\begin{itemize}
    \item \textbf{MCP:} Focused on providing structured context and tool integration for LLMs, emphasizing the connection between models and external resources.
    \item \textbf{ANP:} Introduced decentralized identity mechanisms (e.g., W3C DID), enabling peer-to-peer communication between agents, enhancing system autonomy and flexibility.
    \item \textbf{A2A:} Provided a standardized framework for collaboration between enterprise-level agents, supporting task management, message exchange, and multimodal outputs, thus facilitating cross-platform and multi-vendor agent collaboration.
\end{itemize}
This evolutionary process demonstrates the shift from basic functionality to complex system collaboration, reflecting the continued expansion of the agent ecosystem in terms of scalability and diversity.

Through the aforementioned cases, we can clearly observe the trajectory of development within agent communication protocols and identify potential future iteration targets. While we do not engage in a direct comparison of the protocols' advantages and disadvantages, we offer the following recommendations for agent developers and researchers:
\begin{itemize}
    \item \textbf{Contextual Fit:} Select the appropriate protocol based on the specific application scenario. For example, MCP is ideal for scenarios requiring integration with external tools and data sources; ANP is more suitable for cross-domain communication and collaboration among agents on the Internet; A2A offers more comprehensive support for inter agent collaboration.
    \item \textbf{Focus on Security and Performance:} During the selection and implementation of protocols, attention should be paid to authentication mechanisms, data transmission security, and performance optimization to ensure the reliability and efficiency of the system.
    \item \textbf{Monitor Protocol Evolution:} As agent protocols continue to develop, maintaining awareness of new protocols and versions is essential. Evaluating their impact on existing systems and assessing potential optimization opportunities will be crucial.
\end{itemize}

\section{Use-Case Analysis}
\begin{figure*}[t]
    \centering
    \includegraphics[width=0.99\linewidth]{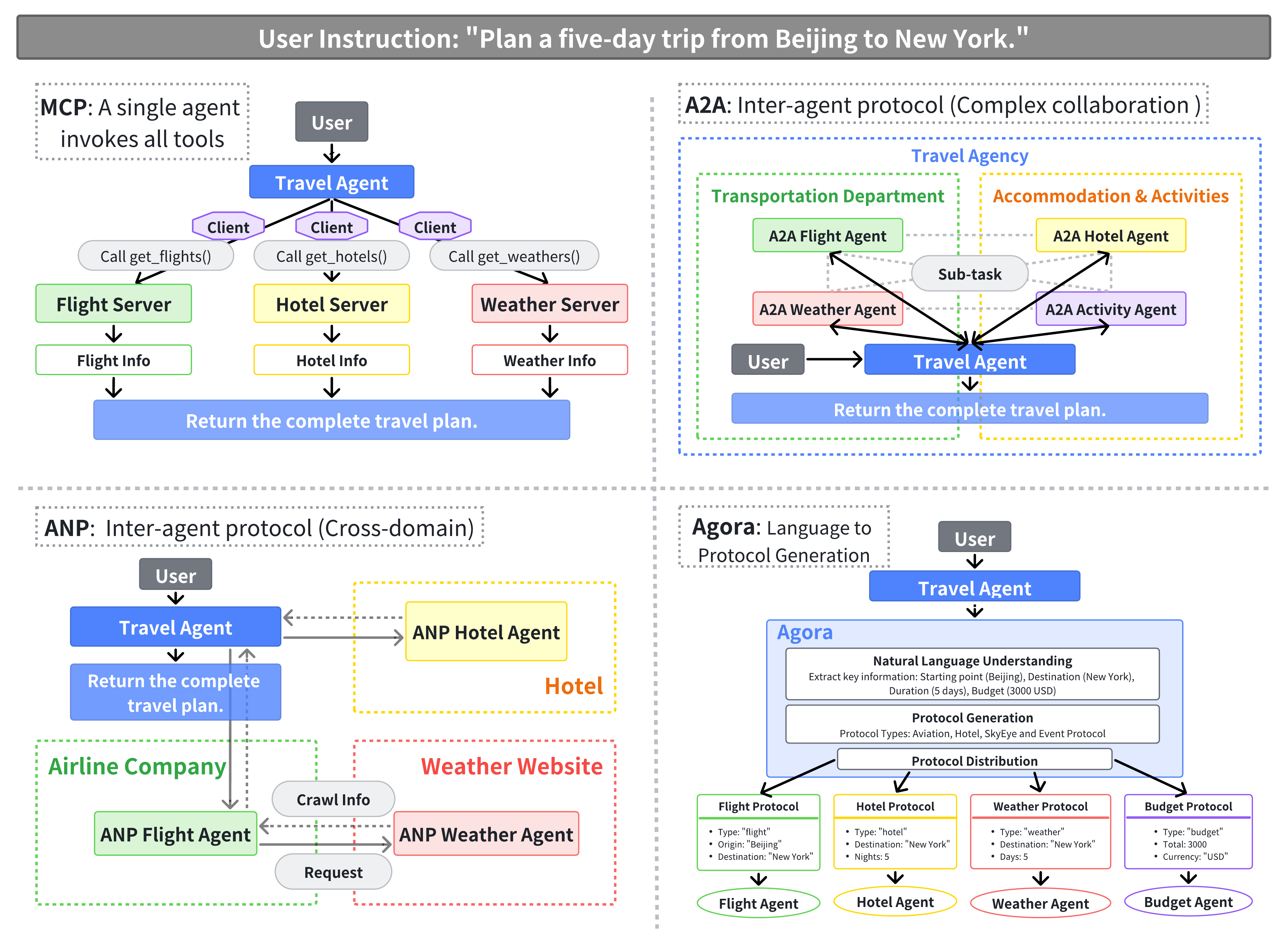}
    \caption{Use-case analyses of four protocols under the same user instruction shown at the top.}
    \label{fig:use-case-analysis}
\end{figure*}

This section provides a comparative analysis of four intelligent agent protocols—MCP, A2A, ANP, and Agora—applied to the same use case: planning a five-day trip from Beijing to New York. Figure \ref{fig:use-case-analysis} illustrates the architectural differences and interaction patterns of each protocol.

\subsection{MCP: Single Agent Invokes All Tools}

Model Context Protocol (MCP) protocol represents a centralized approach where a single agent coordinates all interactions with external services. As shown in Figure \ref{fig:use-case-analysis} (upper left), the MCP Travel Client serves as the coordinating agent with direct dependencies on all external services:
\begin{itemize}
    \item The central MCP Travel Client directly invokes Flight Server, Hotel Server, and Weather Server through respective calls (\texttt{get\_flights()}, \texttt{get\_hotels()}, \texttt{get\_weather()}).
    \item All external services are treated as tools that provide information but don't interact with each other.
    \item Information flow follows a strict star pattern with the MCP Travel Client at the center.
    \item The client is responsible for aggregating all responses and generating the complete travel plan.
\end{itemize}

The MCP architecture excels in simplicity and control but lacks flexibility. The central agent must be aware of all services and their interfaces, creating a high-dependency structure that may be difficult to scale or modify. Additionally, all communication must pass through the central agent, potentially creating a performance bottleneck.

\subsection{A2A: Complex Collaboration Inter-agents Within an Enterprise}

Agent-to-Agent (A2A) protocol enables direct communication between different agents for complex tasks. As depicted in Figure \ref{fig:use-case-analysis} (upper right), the A2A implementation distributes intelligence across multiple specialized agents:
\begin{itemize}
    \item Agents are organized into logical departments (Transportation, Accommodation \& Activities).
    \item Each agent has explicit dependencies: the Flight Agent and Activity Agent depend on the Weather Agent for environmental data.
    \item Agents communicate directly with each other, without requiring central coordination for every interaction.
    \item The A2A Travel Planner functions as a non-central coordinator that primarily collects final results.
\end{itemize}

A2A protocol demonstrates a more flexible and realistic architecture where agents can establish direct connections when needed. For instance, the A2A Flight Agent can obtain weather information directly from the Weather Agent without going through the Travel Planner. This reduces unnecessary communication overhead and allows for more complex collaboration patterns across any type of organization or multi-agent system.

\subsection{ANP: Cross-Domain Agent Protocol}

Agent Network Protocol (ANP) extends collaboration through standardized cross-domain interactions. Figure \ref{fig:use-case-analysis} (lower left) shows how ANP enables negotiations between agents in different organizational domains:
\begin{itemize}
    \item Distinct organizational boundaries separate the Airline Company, Hotel, and Weather Website.
    \item Cross-domain collaboration occurs through formal protocol-based requests and responses.
    \item The Flight Agent negotiates with the Weather Agent across domain boundaries.
    \item The Travel Planner coordinates the overall process but doesn't mediate every interaction.
\end{itemize}

ANP addresses the challenges of collaboration between independent agents by formalizing the protocol-based interaction process. While A2A focuses on message-based delegation, ANP establishes clear protocols for structured requests and responses between agents. This makes it particularly suitable for scenarios involving agents with distinct capabilities, well-defined interfaces, and potentially different security boundaries, regardless of whether they exist within the same system or across multiple systems.

\subsection{Agora: Natural Language to Protocol Generation}

The Agora protocol represents the most user-centric approach, converting natural language requests directly into standardized protocols. As illustrated in Figure \ref{fig:use-case-analysis} (lower right), Agora introduces several distinctive layers:
\begin{itemize}
    \item The process begins with natural language understanding, parsing the user's request into structured components (origin, destination, duration, budget).
    \item The protocol generation layer transforms these components into formalized protocols for different service types.
    \item Protocol distribution dispatches the appropriate protocols to specialized agents (Flight, Hotel, Weather, Budget).
    \item Each agent responds to its specific protocol rather than to free-form requests.
\end{itemize}

Agora's three-stage process (understanding, generation, distribution) creates a highly adaptable system that shields specialized agents from the complexities of natural language processing. This separation of concerns allows domain-specific agents to focus on their core competencies while the Agora layer handles the interpretation of user intent.


Through the case study, it becomes evident that each protocol has specific conditions and dependencies for successful application.  1) MCP employs a centralized agent (e.g., travel assistant) that sequentially invokes tools with clear interfaces to accomplish tasks.  This approach works efficiently for well-defined workflows but may require central agent modification to adapt to new scenarios.  2) A2A enables collaboration through message/data exchange between specialized agents (e.g., flight, hotel, weather).  Each agent autonomously handles its assigned task and communicates results back to a coordinating agent, allowing for flexible communication patterns while maintaining overall coordination.  3) ANP utilizes structured protocol-based interactions where the primary agent retains processing logic but delegates specific execution steps through well-defined API-like interfaces.  This standardized approach works effectively regardless of whether agents exist within the same system or across different domains.  4) Finally, Agora focuses on translating natural language into appropriate structured protocols, serving as an intermediary layer that maps user intentions to the specific protocols required by different agents.  Each protocol's applicability depends on factors such as the desired level of agent autonomy, communication flexibility, interface standardization, and the complexity of the tasks being performed.

\section{Academic Outlook}
The development of agent protocols is progressing rapidly. 
This section outlines the expected evolution of the field in the short, medium, and long term, highlighting research trends, emerging challenges, and forward-looking visions.

\subsection{Short-Term Outlook: From Static to Evolvable}
\paragraph{Evaluation and Benchmarking.}
While various protocols have been proposed for different agent applications, a unified benchmark for evaluating their effectiveness remains less explored. 
Current efforts are converging toward designing evaluation frameworks that go beyond task success, incorporating aspects such as communication efficiency, robustness to environmental changes, adaptability, and scalability. 
The development of diverse simulation environments and standardized testbeds is expected to provide both controlled and open-ended scenarios, thereby facilitating fair and consistent comparisons across protocols.

\paragraph{Privacy-Preserving Protocols.}
With agents increasingly operating in sensitive domains (e.g., healthcare, finance), ensuring secure and confidential communication becomes essential.
Future research should explore the development of protocols that allow agents to exchange information while minimizing the exposure of internal states or personal data. 
Access to information can be managed by authorization mechanisms, potentially based on attributes such as the agent’s role, task, or security clearance, as defined within the communication protocol. 
Additionally, inspired by federated learning~\citep{zhang2021survey}, protocols could facilitate collaboration by enabling agents to share aggregated insights, information derived from locally held private data, or anonymized intermediate results, rather than transmitting raw sensitive information.

\paragraph{Agent Mesh Protocol.}
Existing agent interaction protocols are primarily designed for communication between pairs of agents, which can become increasingly inefficient as the number and complexity of agents grow.
To overcome these limitations, we envision the development of an Agent Mesh Protocol—a communication model inspired by human group chats in the digital age.
This protocol would enable full transparency and shared access to communication history within an agent group, promoting more effective coordination and collaborative decision-making.
Implementing the mesh protocol will require the design of mechanisms that support group-level semantics, maintain consistency and synchronization of shared knowledge, and effectively handle challenges such as message ordering, dynamic group membership, and communication overhead.


\paragraph{Evolvable Protocols.}
Instead of viewing protocols as static rules, future agent systems may incorporate evolvable protocols---treating protocols as dynamic, modular, and learnable components integral to the agents' adaptive capabilities. 
In this paradigm, protocols are not immutable frameworks imposed externally, but rather resources that agents can actively manage and refine. 
Agents could be enabled to retrieve specific protocol modules or combine elements from multiple protocols to construct a customized communication strategy tailored to the requirements of the current task. 
Moreover, agents may be trained to discover optimal protocol variations or negotiation strategies that enhance communication efficiency or increase task success over time. 
This adaptability would allow agent systems to generalize to novel situations, optimize interactions for particular partners or conditions, and potentially scale to more complex collaborative scenarios.

\subsection{Mid-Term Outlook: From Rules to Ecosystems}
\paragraph{Built-In Protocol Knowledge.}
Instead of supplying protocol instructions at inference time, future development may investigate the possibility of training large language models with protocol content and structures integrated into their parameters. 
This allows agents to execute protocol-compliant behavior without explicit prompting, resulting in more efficient and seamless interaction.
However, directly injecting protocol knowledge via training introduces limitations in adaptability---as once a model is trained, it becomes difficult to incorporate updates or modifications to protocol standards. 
Nonetheless, it holds strategic importance for model providers, as the choice of which protocols to embed could influence future standards and competitive dynamics in the agent ecosystem.

\paragraph{Layered Protocol Architectures.}
Protocol design may evolve from the current isolated structure towards layered protocol architectures, which aim to separate concerns across different levels of communication. 
By decoupling low-level transport and synchronization mechanisms from high-level semantic and task-related interactions, such architectures can improve modularity and scalability across heterogeneous agents.. 
Inspired by classical network protocol design, this architecture allows diverse agents to interoperate more efficiently by adhering to shared abstractions at each layer. 
Furthermore, the layered architectures may pave the way for dynamic protocol composition, where agents can negotiate or auto-select interaction layers suited to the context—adapting from rigid rule-following to more fluid, ecosystem-level behavior. 
This adaptability is crucial in mixed human-AI environments, where norms, preferences, and objectives evolve over time. 
Layered protocols could also integrate ethical, legal, and social constraints at higher layers, aligning agent behavior with broader societal values.

\subsection{Long-Term Outlook: From Protocols to Intelligence Infrastructure}

\paragraph{Collective Intelligence and Scaling Laws.}

As agent protocols continue to mature, a compelling long-term direction is to explore the emergence of collective intelligence in large-scale, interconnected agent populations. Building upon prior work in multi-agent systems, swarm intelligence, and complex adaptive networks, future research may investigate the scaling laws of agents and environments—that is, how population size, communication topologies, and protocol configurations jointly shape system-level behaviors, emergent properties, and robustness. Unlike traditional simulations, the advent of internet-native, decentralized agent protocols makes it increasingly feasible to observe and analyze these dynamics at web scale. In the long run, such findings may inform the principled design of distributed agent collectives as a new computational substrate—scalable, adaptive, and capable of exhibiting intelligence beyond individual capabilities.

\paragraph{Agent Data Networks.}

Concurrently, we anticipate the emergence of a dedicated Agent Data Network (ADN)—a foundational data infrastructure optimized for autonomous agent communication and coordination. Unlike traditional web interactions, which are primarily designed for human interpretation and front-end rendering, the ADN would support structured, intent-driven, and protocol-compliant information exchange among agents. While still operating atop the existing internet stack (e.g., TCP/IP and HTTP), the ADN represents a shift in semantic abstraction: agents would increasingly rely on machine-centric data representations, such as latent task states, distributed memory snapshots, and temporal context logs, rather than human-readable web content. This network layer would serve agents' operational needs directly—enabling persistent state synchronization, long-horizon planning, and asynchronous collaboration—without requiring human intervention or visibility.

\section{Conclusion}
In this survey, we provide the first comprehensive analysis of existing AI agent protocols. 
By systematically classifying protocols into two-dimensional classification and evaluating key performance dimensions such as efficiency, scalability, and security, we offer a practical reference for both practitioners and researchers. 
This structured overview not only helps users better navigate the growing ecosystem of agent protocols but also highlights the trade-offs and design considerations involved in building reliable, efficient, and secure agent systems.
Looking ahead, we envision the emergence of next-generation protocols, such as evolvable, privacy-aware, and group-coordinated protocols, as well as the emergence of layered architectures and collective intelligence infrastructures.
The development of agent protocols will pave the way toward a more connected and collaborative agent ecosystemwhere agents and tools can dynamically form coalitions, exchange knowledge, and co-evolve to solve increasingly complex real-world problems. 
Much like the foundational protocols of the internet, future agent communication standards have the potential to unlock a new era of distributed, collective intelligence---reshaping how intelligence is shared, coordinated, and amplified across systems.



\bibliographystyle{unsrtnat}
\bibliography{references}  

\begin{thebibliography}{57}
\providecommand{\natexlab}[1]{#1}
\providecommand{\url}[1]{\texttt{#1}}
\expandafter\ifx\csname urlstyle\endcsname\relax
  \providecommand{\doi}[1]{doi: #1}\else
  \providecommand{\doi}{doi: \begingroup \urlstyle{rm}\Url}\fi

\bibitem[Caffagni et~al.(2024)Caffagni, Cocchi, Barsellotti, Moratelli, Sarto, Baraldi, Cornia, and Cucchiara]{caffagni2024r}
Davide Caffagni, Federico Cocchi, Luca Barsellotti, Nicholas Moratelli, Sara Sarto, Lorenzo Baraldi, Marcella Cornia, and Rita Cucchiara.
\newblock The (r) evolution of multimodal large language models: A survey.
\newblock \emph{arXiv preprint arXiv:2402.12451}, 2024.

\bibitem[OpenAI et~al.(2024)OpenAI, Achiam, Adler, Agarwal, et~al.]{openai2024gpt4technicalreport}
OpenAI, Josh Achiam, Steven Adler, Sandhini Agarwal, et~al.
\newblock Gpt-4 technical report, 2024.
\newblock URL \url{https://arxiv.org/abs/2303.08774}.

\bibitem[Gottweis et~al.(2025)Gottweis, Weng, Daryin, Tu, Palepu, Sirkovic, Myaskovsky, Weissenberger, Rong, Tanno, Saab, Popovici, Blum, Zhang, Chou, Hassidim, Gokturk, Vahdat, Kohli, Matias, Carroll, Kulkarni, Tomasev, Guan, Dhillon, Vaishnav, Lee, Costa, Penadés, Peltz, Xu, Pawlosky, Karthikesalingam, and Natarajan]{gottweis2025aicoscientist}
Juraj Gottweis, Wei-Hung Weng, Alexander Daryin, Tao Tu, Anil Palepu, Petar Sirkovic, Artiom Myaskovsky, Felix Weissenberger, Keran Rong, Ryutaro Tanno, Khaled Saab, Dan Popovici, Jacob Blum, Fan Zhang, Katherine Chou, Avinatan Hassidim, Burak Gokturk, Amin Vahdat, Pushmeet Kohli, Yossi Matias, Andrew Carroll, Kavita Kulkarni, Nenad Tomasev, Yuan Guan, Vikram Dhillon, Eeshit~Dhaval Vaishnav, Byron Lee, Tiago R~D Costa, José~R Penadés, Gary Peltz, Yunhan Xu, Annalisa Pawlosky, Alan Karthikesalingam, and Vivek Natarajan.
\newblock Towards an ai co-scientist, 2025.
\newblock URL \url{https://arxiv.org/abs/2502.18864}.

\bibitem[Yang et~al.(2025{\natexlab{a}})Yang, Huang, Qi, Feng, Hu, Zhu, Hu, Zhao, He, Liu, Wang, Qiu, Cao, Cai, Yu, and Zhang]{yang2025whosmvpgametheoreticevaluation}
Yingxuan Yang, Bo~Huang, Siyuan Qi, Chao Feng, Haoyi Hu, Yuxuan Zhu, Jinbo Hu, Haoran Zhao, Ziyi He, Xiao Liu, Zongyu Wang, Lin Qiu, Xuezhi Cao, Xunliang Cai, Yong Yu, and Weinan Zhang.
\newblock Who's the mvp? a game-theoretic evaluation benchmark for modular attribution in llm agents, 2025{\natexlab{a}}.
\newblock URL \url{https://arxiv.org/abs/2502.00510}.

\bibitem[Guo et~al.(2024)Guo, Chen, Wang, Chang, Pei, Chawla, Wiest, and Zhang]{Guo2024LargeLM}
Taicheng Guo, Xiuying Chen, Yaqi Wang, Ruidi Chang, Shichao Pei, N.~Chawla, Olaf Wiest, and Xiangliang Zhang.
\newblock Large language model based multi-agents: A survey of progress and challenges.
\newblock In \emph{International Joint Conference on Artificial Intelligence}, 2024.
\newblock URL \url{https://api.semanticscholar.org/CorpusID:267412980}.

\bibitem[Zhou et~al.(2024)Zhou, Yang, Wen, Wen, Wang, Xi, Xu, Yu, and Zhang]{zhou2024tradenhancingllmagents}
Ruiwen Zhou, Yingxuan Yang, Muning Wen, Ying Wen, Wenhao Wang, Chunling Xi, Guoqiang Xu, Yong Yu, and Weinan Zhang.
\newblock Trad: Enhancing llm agents with step-wise thought retrieval and aligned decision, 2024.
\newblock URL \url{https://arxiv.org/abs/2403.06221}.

\bibitem[Qu et~al.(2025)Qu, Dai, Wei, Cai, Wang, Yin, Xu, and Wen]{Qu_2025}
Changle Qu, Sunhao Dai, Xiaochi Wei, Hengyi Cai, Shuaiqiang Wang, Dawei Yin, Jun Xu, and Ji-rong Wen.
\newblock Tool learning with large language models: a survey.
\newblock \emph{Frontiers of Computer Science}, 19\penalty0 (8), January 2025.
\newblock ISSN 2095-2236.
\newblock \doi{10.1007/s11704-024-40678-2}.
\newblock URL \url{http://dx.doi.org/10.1007/s11704-024-40678-2}.

\bibitem[Patil et~al.(2023)Patil, Zhang, Wang, and Gonzalez]{patil2023gorillalargelanguagemodel}
Shishir~G. Patil, Tianjun Zhang, Xin Wang, and Joseph~E. Gonzalez.
\newblock Gorilla: Large language model connected with massive apis, 2023.
\newblock URL \url{https://arxiv.org/abs/2305.15334}.

\bibitem[Liu et~al.(2024)Liu, Hoang, Zhang, Zhu, Lan, Kokane, Tan, Yao, Liu, Feng, Murthy, Yang, Savarese, Niebles, Wang, Heinecke, and Xiong]{liu2024apigenautomatedpipelinegenerating}
Zuxin Liu, Thai Hoang, Jianguo Zhang, Ming Zhu, Tian Lan, Shirley Kokane, Juntao Tan, Weiran Yao, Zhiwei Liu, Yihao Feng, Rithesh Murthy, Liangwei Yang, Silvio Savarese, Juan~Carlos Niebles, Huan Wang, Shelby Heinecke, and Caiming Xiong.
\newblock Apigen: Automated pipeline for generating verifiable and diverse function-calling datasets, 2024.
\newblock URL \url{https://arxiv.org/abs/2406.18518}.

\bibitem[Rajaei(2024)]{Multi-Agent-as-a-Service}
Saman Rajaei.
\newblock Multi-agent-as-a-service — a senior engineer’s overview.
\newblock \url{https://medium.com/data-science/multi-agent-as-a-service-a-senior-engineers-overview-fc759f5bbcfa}, 2024.

\bibitem[Yang et~al.(2024)Yang, Peng, Wang, Wen, and Zhang]{yang2024llmbasedmultiagentsystemstechniques}
Yingxuan Yang, Qiuying Peng, Jun Wang, Ying Wen, and Weinan Zhang.
\newblock Llm-based multi-agent systems: Techniques and business perspectives, 2024.
\newblock URL \url{https://arxiv.org/abs/2411.14033}.

\bibitem[Chen et~al.(2024)Chen, You, Li, Guan, Qian, Zhao, Yang, Xie, Liu, and Sun]{chen2024internet}
Weize Chen, Ziming You, Ran Li, Yitong Guan, Chen Qian, Chenyang Zhao, Cheng Yang, Ruobing Xie, Zhiyuan Liu, and Maosong Sun.
\newblock Internet of agents: Weaving a web of heterogeneous agents for collaborative intelligence.
\newblock \emph{arXiv preprint arXiv:2407.07061}, 2024.

\bibitem[Yang et~al.(2025{\natexlab{b}})Yang, Chai, Shao, Song, Qi, Rui, and Zhang]{yang2025agentnetdecentralizedevolutionarycoordination}
Yingxuan Yang, Huacan Chai, Shuai Shao, Yuanyi Song, Siyuan Qi, Renting Rui, and Weinan Zhang.
\newblock Agentnet: Decentralized evolutionary coordination for llm-based multi-agent systems, 2025{\natexlab{b}}.
\newblock URL \url{https://arxiv.org/abs/2504.00587}.

\bibitem[Anthropic(2024)]{anthropic2025}
Anthropic.
\newblock Model context protocol, 2024.
\newblock URL \url{https://www.anthropic.com/news/model-context-protocol}.
\newblock Accessed: 2025-04-19.

\bibitem[Chang(2024)]{anp2024}
Gaowei Chang.
\newblock Anp: Agent network protocol, 2024.
\newblock URL \url{https://www.agent-network-protocol.com/}.
\newblock Accessed: 2025-04-21.

\bibitem[Google(2025)]{a2a2025}
Google.
\newblock A2a: Agent2agent protocol, 2025.
\newblock URL \url{https://github.com/google/A2A}.
\newblock Accessed: 2025-04-21.

\bibitem[Yao et~al.(2022)]{yao2022react}
Shinnung Yao et~al.
\newblock React: Synergizing reasoning and acting in language models.
\newblock \emph{arXiv preprint arXiv:2210.03629}, 2022.

\bibitem[Tang et~al.(2023)Tang, Li, Chen, Lin, and Zhang]{autogpt}
Tao Tang, Zhihui Li, Jiangjie Chen, Mingyu Lin, and Wei Zhang.
\newblock Autogpt: An autonomous gpt-4 experiment.
\newblock \emph{arXiv preprint arXiv:2308.08155}, 2023.

\bibitem[Hong et~al.(2024)Hong, Wang, Yang, Guo, Chen, and Li]{hong2024metagpt}
Sirui Hong, Xiawu Wang, Mingyu Yang, Jiale Guo, Di~Chen, and Bingchen Li.
\newblock Metagpt: Meta programming for multi-agent collaborative framework.
\newblock \emph{arXiv preprint arXiv:2401.03066}, 2024.

\bibitem[Zhao et~al.(2025)Zhao, Zhou, Li, Tang, Wang, Hou, Min, Zhang, Zhang, Dong, Du, Yang, Chen, Chen, Jiang, Ren, Li, Tang, Liu, Liu, Nie, and Wen]{zhao2025surveylargelanguagemodels}
Wayne~Xin Zhao, Kun Zhou, Junyi Li, Tianyi Tang, Xiaolei Wang, Yupeng Hou, Yingqian Min, Beichen Zhang, Junjie Zhang, Zican Dong, Yifan Du, Chen Yang, Yushuo Chen, Zhipeng Chen, Jinhao Jiang, Ruiyang Ren, Yifan Li, Xinyu Tang, Zikang Liu, Peiyu Liu, Jian-Yun Nie, and Ji-Rong Wen.
\newblock A survey of large language models, 2025.
\newblock URL \url{https://arxiv.org/abs/2303.18223}.

\bibitem[Yin et~al.(2024)Yin, Fu, Zhao, Li, Sun, Xu, and Chen]{Yin_2024}
Shukang Yin, Chaoyou Fu, Sirui Zhao, Ke~Li, Xing Sun, Tong Xu, and Enhong Chen.
\newblock A survey on multimodal large language models.
\newblock \emph{National Science Review}, 11\penalty0 (12), November 2024.
\newblock ISSN 2053-714X.
\newblock \doi{10.1093/nsr/nwae403}.
\newblock URL \url{http://dx.doi.org/10.1093/nsr/nwae403}.

\bibitem[Zhang et~al.(2024)Zhang, Bo, Ma, Li, Chen, Dai, Zhu, Dong, and Wen]{zhang2024surveymemorymechanismlarge}
Zeyu Zhang, Xiaohe Bo, Chen Ma, Rui Li, Xu~Chen, Quanyu Dai, Jieming Zhu, Zhenhua Dong, and Ji-Rong Wen.
\newblock A survey on the memory mechanism of large language model based agents, 2024.
\newblock URL \url{https://arxiv.org/abs/2404.13501}.

\bibitem[Wang et~al.(2023)Wang, Qin, Li, Zhang, Li, et~al.]{wang2023toolllm}
Yujia Wang, Yusheng Qin, Haozhe Li, Guan Zhang, Xin Li, et~al.
\newblock Toolllm: Facilitating large language models to master 16000+ real-world apis.
\newblock \emph{arXiv preprint arXiv:2307.16789}, 2023.

\bibitem[Schick et~al.(2023)Schick, Dwivedi-Yu, Dessì, Raileanu, Lomeli, Zettlemoyer, Cancedda, and Scialom]{schick2023toolformerlanguagemodelsteach}
Timo Schick, Jane Dwivedi-Yu, Roberto Dessì, Roberta Raileanu, Maria Lomeli, Luke Zettlemoyer, Nicola Cancedda, and Thomas Scialom.
\newblock Toolformer: Language models can teach themselves to use tools, 2023.
\newblock URL \url{https://arxiv.org/abs/2302.04761}.

\bibitem[Liu et~al.(2023)Liu, Zhou, Zhang, Peng, et~al.]{liu2023agentbench}
Xiao Liu, Hao Zhou, Zhiheng Zhang, Dian Peng, et~al.
\newblock Agentbench: Evaluating llms as agents.
\newblock \emph{arXiv preprint arXiv:2308.03688}, 2023.

\bibitem[{VentureBeat}(2024)]{microsoft2024agents}
{VentureBeat}.
\newblock Microsoft's 10 new ai agents strengthen its enterprise automation lead.
\newblock \url{https://venturebeat.com/ai/microsofts-10-new-ai-agents-strengthen-its-enterprise-automation-lead/}, 2024.
\newblock Accessed: 2024-04-23.

\bibitem[{IBM Newsroom}(2024)]{ibm2024agent}
{IBM Newsroom}.
\newblock Ibm introduces new ai integration services to help enterprises build and scale ai.
\newblock \url{https://newsroom.ibm.com/blog-ibm-introduces-new-ai-integration-services-to-help-enterprises-build-and-scale-ai}, 2024.
\newblock Accessed: 2024-04-23.

\bibitem[{TrustedBy.ai}(2024)]{trustedby2024coze}
{TrustedBy.ai}.
\newblock Comparing 9 ai agent development platforms: Dify, coze, adept, kognitos, flowise, articul8, stack ai.
\newblock \url{https://trustedby.ai/blog/comparing-9-ai-agent-development-platforms-dify-coze-adept-kognitos-flowise-articul8-stack-ai}, 2024.
\newblock Accessed: 2024-04-23.

\bibitem[Jaech et~al.(2024)Jaech, Kalai, Lerer, Richardson, El-Kishky, Low, Helyar, Madry, Beutel, Carney, et~al.]{jaech2024openai}
Aaron Jaech, Adam Kalai, Adam Lerer, Adam Richardson, Ahmed El-Kishky, Aiden Low, Alec Helyar, Aleksander Madry, Alex Beutel, Alex Carney, et~al.
\newblock Openai o1 system card.
\newblock \emph{arXiv preprint arXiv:2412.16720}, 2024.

\bibitem[{LangChain}(2024)]{langchain2024langgraph}
{LangChain}.
\newblock Langgraph: Building graph-based agent workflows.
\newblock \url{https://www.langchain.com/langgraph}, 2024.
\newblock Accessed: 2024-04-23.

\bibitem[{Microsoft Learn}(2024)]{microsoft2024semantic}
{Microsoft Learn}.
\newblock Semantic kernel agent framework.
\newblock \url{https://learn.microsoft.coralm/en-us/semantic-kernel/frameworks/agent/}, 2024.
\newblock Accessed: 2024-04-23.

\bibitem[Liu et~al.(2025)Liu, Li, Zhang, Wang, He, Hong, Liu, Zhang, Song, Zhu, Cheng, Wang, Wang, Luo, Jin, Zhang, Liu, Chen, Zhang, Yu, Shi, Li, Wu, Teng, Jia, Xu, Xiang, Lin, Liu, Liu, Su, Sun, Berseth, Nie, Foster, Ward, Wu, Gu, Zhuge, Tang, Wang, You, Wang, Pei, Yang, Qi, and Wu]{liu2025advanceschallengesfoundationagents}
Bang Liu, Xinfeng Li, Jiayi Zhang, Jinlin Wang, Tanjin He, Sirui Hong, Hongzhang Liu, Shaokun Zhang, Kaitao Song, Kunlun Zhu, Yuheng Cheng, Suyuchen Wang, Xiaoqiang Wang, Yuyu Luo, Haibo Jin, Peiyan Zhang, Ollie Liu, Jiaqi Chen, Huan Zhang, Zhaoyang Yu, Haochen Shi, Boyan Li, Dekun Wu, Fengwei Teng, Xiaojun Jia, Jiawei Xu, Jinyu Xiang, Yizhang Lin, Tianming Liu, Tongliang Liu, Yu~Su, Huan Sun, Glen Berseth, Jianyun Nie, Ian Foster, Logan Ward, Qingyun Wu, Yu~Gu, Mingchen Zhuge, Xiangru Tang, Haohan Wang, Jiaxuan You, Chi Wang, Jian Pei, Qiang Yang, Xiaoliang Qi, and Chenglin Wu.
\newblock Advances and challenges in foundation agents: From brain-inspired intelligence to evolutionary, collaborative, and safe systems, 2025.
\newblock URL \url{https://arxiv.org/abs/2504.01990}.

\bibitem[WildCardAI(2025)]{agentsjson2025}
WildCardAI.
\newblock agents.json specification.
\newblock \url{https://github.com/wild-card-ai/agents-json}, 2025.
\newblock Accessed: 2025-04-22.

\bibitem[NEAR(2025)]{aitp}
NEAR.
\newblock Aitp: Agent interaction \& transaction protocol, 2025.
\newblock URL \url{https://aitp.dev/}.
\newblock Accessed: 2025-04-22.

\bibitem[AI and Data(2025)]{agentcommunicationptl}
Linux~Foundation AI and IBM Data.
\newblock Acp: Agent communication protocol, 2025.
\newblock URL \url{https://agentcommunicationprotocol.dev/introduction/welcome}.
\newblock Accessed: 2025-05-28.

\bibitem[Cisco(2025)]{agentconnectptl}
Galileo Cisco, Langchain.
\newblock Acp: Agent connect protocol, 2025.
\newblock URL \url{https://spec.acp.agntcy.org/}.
\newblock Accessed: 2025-04-22.

\bibitem[Marro et~al.(2024)Marro, Malfa, Wright, Li, Shadbolt, Wooldridge, and Torr]{marro2024scalablecommunicationprotocolnetworks}
Samuele Marro, Emanuele~La Malfa, Jesse Wright, Guohao Li, Nigel Shadbolt, Michael Wooldridge, and Philip Torr.
\newblock A scalable communication protocol for networks of large language models, 2024.
\newblock URL \url{https://arxiv.org/abs/2410.11905}.

\bibitem[Eclipse(2025)]{lmos2025}
Eclipse.
\newblock Language model operating system (lmos).
\newblock \url{https://eclipse.dev/lmos/}, 2025.
\newblock Accessed: 2025-04-22.

\bibitem[AlEngineerFoundation(2025)]{agentprotocol2025}
AlEngineerFoundation.
\newblock Agent protocol.
\newblock \url{https://agentprotocol.ai/}, 2025.
\newblock Accessed: 2025-04-22.

\bibitem[Ranjan et~al.(2025)Ranjan, Gupta, and Singh]{ranjan2025lokaprotocoldecentralizedframework}
Rajesh Ranjan, Shailja Gupta, and Surya~Narayan Singh.
\newblock Loka protocol: A decentralized framework for trustworthy and ethical ai agent ecosystems, 2025.
\newblock URL \url{https://arxiv.org/abs/2504.10915}.

\bibitem[Srinivasan et~al.(2024)Srinivasan, Bania, V, Mestha, and Liu]{srinivasan2024implementationapplicationintelligibilityprotocol}
Ashwin Srinivasan, Karan Bania, Shreyas V, Harshvardhan Mestha, and Sidong Liu.
\newblock Implementation and application of an intelligibility protocol for interaction with an llm, 2024.
\newblock URL \url{https://arxiv.org/abs/2410.20600}.

\bibitem[Bae et~al.(2025)Bae, Lee, and Jeon]{bae2025continuouslocomotivecrowdbehavior}
Inhwan Bae, Junoh Lee, and Hae-Gon Jeon.
\newblock Continuous locomotive crowd behavior generation, 2025.
\newblock URL \url{https://arxiv.org/abs/2504.04756}.

\bibitem[Gąsieniec et~al.(2024)Gąsieniec, Łukasz Kuszner, Latif, Parasuraman, Spirakis, and Stachowiak]{gąsieniec2024anonymousdistributedlocalisationspatial}
Leszek Gąsieniec, Łukasz Kuszner, Ehsan Latif, Ramviyas Parasuraman, Paul Spirakis, and Grzegorz Stachowiak.
\newblock Anonymous distributed localisation via spatial population protocols, 2024.
\newblock URL \url{https://arxiv.org/abs/2411.08434}.

\bibitem[Stone and Veloso(2000)]{stone2000multiagent}
Peter Stone and Manuela Veloso.
\newblock Multiagent systems: A survey from a machine learning perspective.
\newblock \emph{Autonomous Robots}, 8:\penalty0 345--383, 2000.

\bibitem[Dorri et~al.(2018)Dorri, Kanhere, and Jurdak]{dorri2018multi}
Ali Dorri, Salil~S Kanhere, and Raja Jurdak.
\newblock Multi-agent systems: A survey.
\newblock \emph{Ieee Access}, 6:\penalty0 28573--28593, 2018.

\bibitem[Georgio et~al.(2025)Georgio, Forder, Deb, Carroll, and G{\"u}rcan]{agentcoralprotocol}
Roman~J Georgio, Caelum Forder, Suman Deb, Peter Carroll, and {\"O}nder G{\"u}rcan.
\newblock The coral protocol, 2025.
\newblock URL \url{https://docs.coralprotocol.org/CoralDoc/Introduction/WhatisCoralProtocol}.
\newblock Accessed: 2025-05-28.

\bibitem[Gilbert(2019)]{gilbert2019agent}
Nigel Gilbert.
\newblock \emph{Agent-based models}.
\newblock Sage Publications, 2019.

\bibitem[Wei et~al.(2022)Wei, Wang, Schuurmans, Bosma, Xia, Chi, Le, Zhou, et~al.]{wei2022chain}
Jason Wei, Xuezhi Wang, Dale Schuurmans, Maarten Bosma, Fei Xia, Ed~Chi, Quoc~V Le, Denny Zhou, et~al.
\newblock Chain-of-thought prompting elicits reasoning in large language models.
\newblock \emph{Advances in neural information processing systems}, 35:\penalty0 24824--24837, 2022.

\bibitem[Collins et~al.(2022)Collins, Wong, Feng, Wei, and Tenenbaum]{collins2022structuredflexiblerobustbenchmarking}
Katherine~M. Collins, Catherine Wong, Jiahai Feng, Megan Wei, and Joshua~B. Tenenbaum.
\newblock Structured, flexible, and robust: benchmarking and improving large language models towards more human-like behavior in out-of-distribution reasoning tasks, 2022.
\newblock URL \url{https://arxiv.org/abs/2205.05718}.

\bibitem[Pyatkin et~al.(2022)Pyatkin, Hwang, Srikumar, Lu, Jiang, Choi, and Bhagavatula]{pyatkin2022reinforced}
Valentina Pyatkin, Jena~D Hwang, Vivek Srikumar, Ximing Lu, Liwei Jiang, Yejin Choi, and Chandra Bhagavatula.
\newblock Reinforced clarification question generation with defeasibility rewards for disambiguating social and moral situations.
\newblock \emph{arXiv preprint arXiv:2212.10409}, 2022.

\bibitem[Zhong and Wang(2023)]{zhong2023study}
Li~Zhong and Zilong Wang.
\newblock A study on robustness and reliability of large language model code generation.
\newblock \emph{arXiv preprint arXiv:2308.10335}, 2, 2023.

\bibitem[Hu et~al.(2022)Hu, Shen, Wallis, Allen-Zhu, Li, Wang, Wang, Chen, et~al.]{hu2022lora}
Edward~J Hu, Yelong Shen, Phillip Wallis, Zeyuan Allen-Zhu, Yuanzhi Li, Shean Wang, Lu~Wang, Weizhu Chen, et~al.
\newblock Lora: Low-rank adaptation of large language models.
\newblock \emph{ICLR}, 1\penalty0 (2):\penalty0 3, 2022.

\bibitem[{OTA Tech AI}(2025)]{ota_webagent_2025}
{OTA Tech AI}.
\newblock {Web Agent Protocol}.
\newblock \url{https://github.com/OTA-Tech-AI/web-agent-protocol}, 2025.
\newblock Accessed: 2025-06-11.

\bibitem[Liu et~al.(2022)Liu, Dou, Zhang, Zhang, and Zong]{liu2022multi}
Da~Liu, Liqian Dou, Ruilong Zhang, Xiuyun Zhang, and Qun Zong.
\newblock Multi-agent reinforcement learning-based coordinated dynamic task allocation for heterogenous uavs.
\newblock \emph{IEEE Transactions on Vehicular Technology}, 72\penalty0 (4):\penalty0 4372--4383, 2022.

\bibitem[Jiang et~al.(2018)Jiang, Ghadikolaei, Fodor, Modiano, Pang, Zorzi, and Fischione]{jiang2018lowlatencynetworkinglatencylurks}
Xiaolin Jiang, Hossein~S. Ghadikolaei, Gabor Fodor, Eytan Modiano, Zhibo Pang, Michele Zorzi, and Carlo Fischione.
\newblock Low-latency networking: Where latency lurks and how to tame it, 2018.
\newblock URL \url{https://arxiv.org/abs/1808.02079}.

\bibitem[Fuller and Li(1993)]{fuller1993classless}
V.~Fuller and T.~Li.
\newblock Classless inter-domain routing (cidr): an address assignment and aggregation strategy.
\newblock Technical report, 1993.

\bibitem[Zhang et~al.(2021)Zhang, Xie, Bai, Yu, Li, and Gao]{zhang2021survey}
Chen Zhang, Yu~Xie, Hang Bai, Bin Yu, Weihong Li, and Yuan Gao.
\newblock A survey on federated learning.
\newblock \emph{Knowledge-Based Systems}, 216:\penalty0 106775, 2021.

\end{thebibliography}






\end{document}